\pdfoutput=1
\documentclass[journal]{IEEEtran}

\ifCLASSINFOpdf
\else
   \usepackage[dvips]{graphicx}
\fi
\usepackage{url}

\newif\ifarxiv
\arxivtrue

\usepackage{graphicx}

\usepackage{siunitx}
\usepackage{adjustbox}
\usepackage{booktabs}
\usepackage{multirow}
\usepackage{subcaption}
\usepackage{multirow}
\usepackage{makecell}
\usepackage{pifont}
\usepackage{amsmath}

\usepackage{etoolbox}
\usepackage{cite}

\def\MYTITLE{Fourier-based Action Recognition for Wildlife Behavior Quantification with Event Cameras}

\usepackage{xcolor}
\definecolor{eccvblue}{rgb}{0.12,0.49,0.85}
\usepackage[breaklinks=true,
    colorlinks=true,
    citecolor=eccvblue,
    bookmarksdepth=2,
    bookmarksopen=true,
    bookmarksopenlevel=2
    ]{hyperref}
\hypersetup{
  pdftitle={\MYTITLE},
  pdfsubject={Computer Vision, Pattern Recognition},
  pdfauthor={Friedhelm Hamann, Suman Ghosh, Ignacio Juarez Martinez, Tom Hart, Alex Kacelnik, Guillermo Gallego},
  pdfkeywords={Event Cameras, Intelligent Systems, Penguins, Remote Behavioral Monitoring}
}

\usepackage[capitalize]{cleveref}
\newif\ifaisy
\aisyfalse
\ifaisy
\crefname{section}{Section}{Sections}
\crefname{table}{Table}{Tables}
\crefname{figure}{Figure}{Figures}
\else 
\crefname{section}{Sec.}{Secs.}
\crefname{table}{Tab.}{Tabs.} 
\crefname{figure}{Fig.}{Figs.}
\fi
\Crefname{section}{Section}{Sections}
\Crefname{table}{Table}{Tables}
\Crefname{figure}{Figure}{Figures}

\def\pol{p} 

\def\cE{\mathcal{E}} 

\def\flower{f_l}
\def\fupper{f_u}
\def\fmid{f_{\text{mid}}}

\newcommand{\cmark}{\ding{51}}%
\newcommand{\xmark}{\ding{55}}%

\usepackage{xcolor}
\definecolor{light-gray}{gray}{0.6}
\newcommand\gframe[1]{{\color{light-gray}\frame{#1}}}

\ifarxiv
\usepackage[absolute]{textpos}
\fi

\begin{document}
\title{\MYTITLE}

\ifarxiv
\definecolor{somegray}{gray}{0.5}
\newcommand{\darkgrayed}[1]{\textcolor{somegray}{#1}}
\begin{textblock}{11}(2.5, 0.4)  
\begin{center}
\darkgrayed{This journal paper has been accepted for publication at Advanced Intelligent Systems, 2024.}
\end{center}
\end{textblock}
\fi

\author{Friedhelm Hamann$^{1,2}$, Suman Ghosh$^{1}$, Ignacio Ju\'arez Mart\'inez$^{3}$,\\ Tom Hart$^{4}$, Alex Kacelnik$^{2,3}$, Guillermo Gallego$^{1,2,5}$
\thanks{$^{1}$~Department of Electrical Engineering and Computer Science, Technische Universit\"at Berlin, Berlin, 
and Robotics Institute Germany, Berlin, Germany.
$^{2}$~Science of Intelligence Excellence Cluster, Berlin, Germany.
$^{3}$~Department of Biology, University of Oxford, Oxford, UK.
$^{4}$~Oxford Brookes University, Oxford, UK.
$^{5}$~Einstein Center Digital Future, Berlin, Germany. 
}
\thanks{\url{https://tub-rip.github.io/eventpenguins/}}
}
\maketitle

\begin{abstract}
Event cameras are novel bio-inspired vision sensors that measure pixel-wise brightness changes asynchronously instead of images at a given frame rate.
They offer promising advantages, namely a high dynamic range, low latency, and minimal motion blur.
Modern computer vision algorithms often rely on artificial neural network approaches, which require image-like representations of the data and cannot fully exploit the characteristics of event data.
We propose approaches to action recognition based on the Fourier Transform.
The approaches are intended to recognize oscillating motion patterns commonly present in nature. 
In particular, we apply our approaches to a recent dataset of breeding penguins annotated for ``ecstatic display'', a behavior where the observed penguins flap their wings at a certain frequency.
We find that our approaches are both simple and effective, producing slightly lower results than a deep neural network (DNN) while relying just on a tiny fraction of the parameters compared to the DNN (five orders of magnitude fewer parameters).
They work well despite the uncontrolled, diverse data present in the dataset. 
We hope this work opens a new perspective on event-based processing and action recognition.
\end{abstract}


\IEEEpeerreviewmaketitle

\section{Introduction}
\label{sec:intro}

Inspired by biological processes in the human eye, event cameras have emerged as a promising alternative to conventional cameras \cite{Posch14ieee}.
These novel sensors record pixel-wise brightness changes instead of intensity images (frames), offering significant advantages such as low latency, low power, and high dynamic range. 
However, state-of-the-art approaches for high-level vision tasks, such as object/action recognition, primarily rely on image-based deep neural networks, which require large amounts of training data and computation, thus having limited applicability in low-power scenarios.
Consequently, there is a growing need for more efficient approaches that leverage the unique characteristics of event data~\cite{Gallego20pami}.

\begin{figure}[t]
   \centering
   \includegraphics[trim={8.5cm 6.5cm 8cm 5.5cm},clip,width=\linewidth]{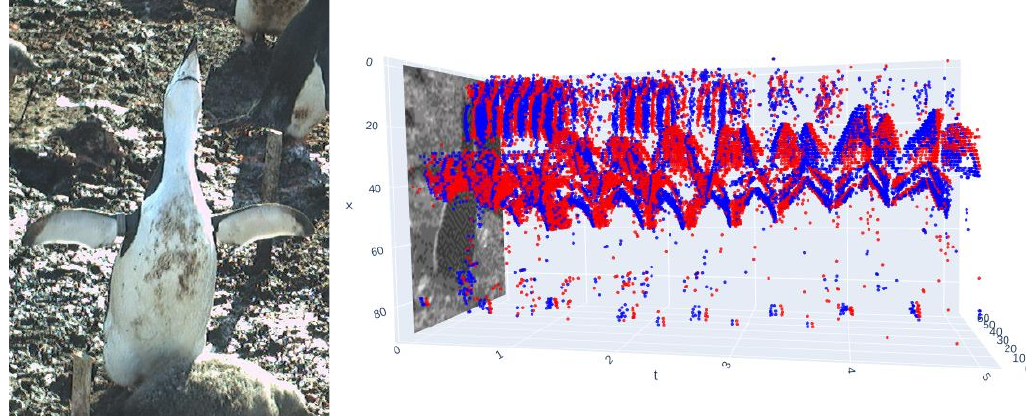}
\caption{\emph{Using Fourier Analysis for action recognition with event cameras}. 
The figure shows a picture of a penguin flapping its wings (Left) 
and the corresponding behavior acquired by an event camera (Right).
The rate at which the event data is produced by the penguin shows a clear oscillatory character.
We leverage this observation to build simple and effective classifiers in the Fourier domain of the event data.
\vspace{-2.24ex}}
\label{fig:intro:eyecatcher}
\end{figure}

This paper presents novel methods for action recognition with event cameras based on the fast Fourier Transform (FFT).
We demonstrate them on a recent dataset of breeding penguins \cite{Hamann24cvpr}, successfully detecting a behavior known as ``ecstatic display'' (ED).
This behavior is characterized by a nesting penguin standing upright, pointing its head upwards, and energetically beating its wings. 
It comprises a dominant oscillatory motion that is naturally recorded by patterns of event data in space-time (\cref{fig:intro:eyecatcher}).
We identify this behavior by its frequency signature using Fourier analysis. 
The task is challenging because there are similar behaviors, like casual wing flaps, that are not ED 
and because the dataset contains natural, uncontrolled conditions, namely varying illumination, snow and rain.
We formulate the task as a classification one. 
The input is the event data of one penguin nest during a certain time window (e.g., 5s), 
and the output is the classification result (ED or background).

The key ideas are:
($i$) event cameras naturally respond to motion in the scene, hence they provide a strong response when penguins perform ED and other motions, 
and ($ii$) the distinguishable response can be simply summarized into a low-dimensional signal or feature that captures the desired oscillatory motion, which simplifies the subsequent classification task using Fourier analysis.
We explore three classifiers on the Fourier spectrum of the signal, which are physically guided by domain knowledge or utilize generic learning-based methods (i.e., artificial neural networks -- ANNs).
The simplest solution consists of a decision threshold on the energy band around the flapping frequency of the ED, which agrees with human intuition. 
The experiments show moderately robust classification, with an average F1 score of 0.54, 
despite the above-mentioned challenges of the dataset \cite{Hamann24cvpr}.
The advantages of event cameras exploited in this application are: 
the high dynamic range (HDR), low-power consumption, and continuous recording in response to motion while suppressing temporal redundancy; specifically \cite{Hamann24cvpr} report 3$\times$ lower power consumption and 10$\times$ lower storage compared to a frame-based camera at an equivalent rate.
To demonstrate the high-speed capabilities of the camera, we carry out experiments in other applications related to industrial inspection (\cref{sec:discussion}).

Our contributions are:
\begin{itemize}
    \item A new action recognition pipeline for the detection of oscillating processes in animal behavior
    based on the FFT of a summarizing event signal (\cref{sec:method}).
    \item Three classification methods on the Fourier-transformed data (\cref{sec:method:energy_band,sec:method:spectrumclassifiers}).
    \item A thorough evaluation of the proposed methods in comparison with four other approaches in terms of classification accuracy and number of model parameters 
    (including sensitivities studies on the parameters and variations of the estimation methods) 
    on the first and challenging event camera dataset recorded in Antarctica (\cref{sec:exp}).
    \item The potential applicability of the method to vibration monitoring scenarios, including high-frequency phenomena beyond the rates of conventional cameras (\cref{sec:discussion}).
\end{itemize}

We believe our work contributes to the field by introducing novel and efficient methods for action recognition using event cameras, fostering research at the intersection of biology and computer vision.
Furthermore, we hope that our study inspires new ideas for efficient action recognition, paving the way for the development of real-time and low-power applications that take full advantage of the unique properties of event cameras.

\section{Related Work}
\label{sec:related_work}

\begin{figure*}
    \centering
    \includegraphics[trim={0.5cm 8.3cm 0cm 1.8cm},clip, width=0.95\linewidth]{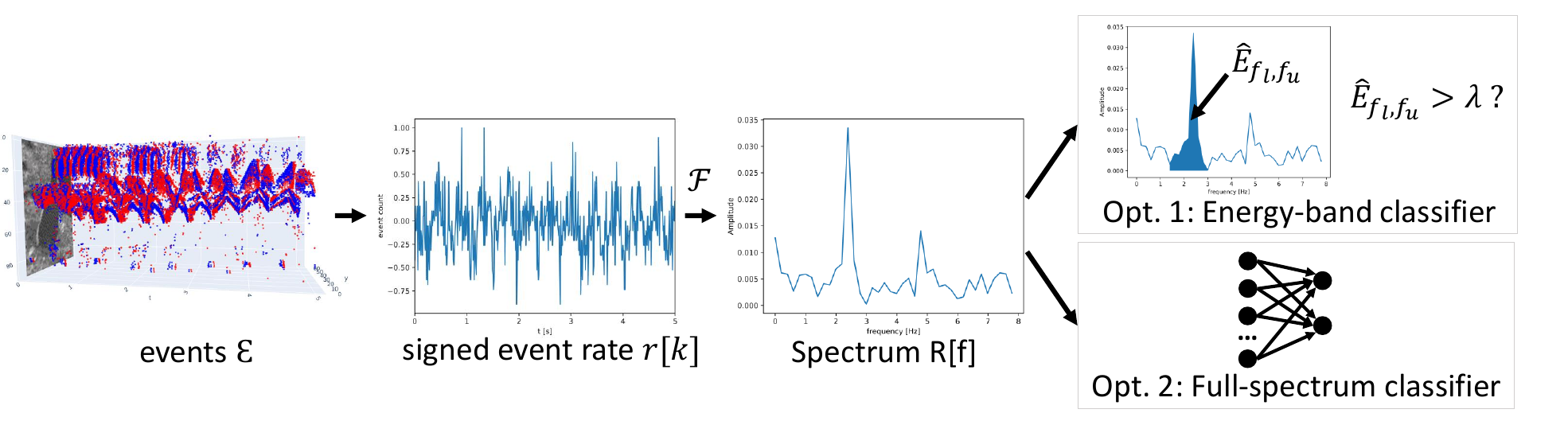}
    \caption{\emph{System overview}.
    The event data in a time window is summarized into a lower dimensional signal (signed event rate) $r[k]$,
    which is transformed into the Fourier domain, $R[f]$.
    Classification criteria can be established based on the particular properties of $R[f]$.
    For example, if the energy in a specific frequency band is higher than a threshold, we assume an ecstatic display is present in the data (Option 1). Likewise, it is possible to train a neural network on the classification task (Option 2).}    
    \label{fig:method:method}
\end{figure*}

\subsubsection{Action Recognition}
\label{sec:related_work:action}
Action Recognition is a vital research area in computer vision, aimed at understanding and classifying human and animal actions from image or video data.
With the rise of deep learning, convolutional neural networks (CNNs) have become dominant in action recognition tasks, achieving state-of-the-art results \cite{Wang16eccv}.
Researchers have explored 3D CNNs \cite{Tran18cvpr,Carreira17cvpr} and two-stream CNN architectures \cite{Wang16eccv,Yang20cvpr,Simonyan14neurips} to better capture spatial and temporal information in videos.
Recently, transformer-based architectures have been introduced \cite{Bertasius21icml} for action recognition.
The trend of recent action recognition methods is towards more complex learned models with a large amount of parameters.
These model types typically require high computational resources, large datasets, and are prohibitively expensive for online applications.

\subsubsection{Event cameras} 
Event cameras operate fundamentally different from conventional cameras \cite{Posch14ieee}. 
They have been used for different applications in computer vision and robotics, like 3D reconstruction \cite{Rebecq18ijcv,Ghosh22aisy}, optical flow estimation \cite{Shiba22eccv,Shiba22aisy} or tracking \cite{Gehrig19ijcv,Hamann22icprvaib}.
With the emergence of event cameras new research fields like image reconstruction \cite{Munda18ijcv,Mostafavi21ijcv} and event-based image deblurring \cite{Zhou2023ijcv} have been developed.
A comprehensive overview can be found in \cite{Gallego20pami}.
Event cameras have gained attention in action recognition tasks due to their natural response to motion and their high dynamic range.
Several datasets exist for the task. 
For example, using a data conversion approach introduced by \cite{Orchard15fns}, the popular UCF-50 Action Recognition Dataset \cite{Reddy13mva} became event-based in \cite{Hu16fns}.
Other datasets are available for gesture and sign language recognition \cite{Amir17cvpr,Vasudevan22paa,Bi19iccv}.

Event-based action recognition has been explored through various methods.
Some try to leverage the event-data sparsity by using graph neural networks \cite{Bi19iccv,Bi20tip,Schaefer22cvpr,Gehrig22arxiv} or PointNet-like architectures \cite{Wang19wcav,Sekikawa19cvpr}.
Others use Spiking Neural Networks (SNNs) to solve the task \cite{Wu18fns,Shrestha18nips,Kaiser20fns}.
More popular and straightforward approaches convert the data into some tensor-like representation (e.g., 2D histograms \cite{Maqueda18cvpr} or time maps \cite{Lagorce17pami}) and subsequently feed it to a regular (i.e., non-spiking) network, such as a CNN \cite{Maro20fns}.

\subsubsection{Wildlife Animal Observation}
Recently, computer vision has emerged as a valuable tool for wildlife animal observation and monitoring \cite{Pereira20nature,Weinstein18jae,Kay22eccv}.
Research efforts have developed and improved techniques for detecting \cite{Beery18eccv,Bondi20wacv}, tracking \cite{Kay22eccv,Bondi20wacv,Wang2023ijcv,Zhang2023ijcv,Xiao2023ijcv}, and recognizing \cite{Beery18cvprw,Berg14cvpr} animals and their behaviors in their natural habitats.
The works in \cite{Patel2023ijcv,Li2023ijcv,Yao2023ijcv} apply computer vision methods for animal pose estimation and tracking.

Early applications of computer vision in wildlife observation involve the use of camera traps to capture images or videos of animals as they pass by \cite{Sollmann18aje}.
These systems have been utilized for monitoring species abundance, distribution and activity patterns.
Machine learning techniques, particularly deep learning (DL) methods, have been employed to automate the analysis of camera trap images.
For instance, \cite{Norouzzadeh18pnas} demonstrated the use of DL for automatically identifying and classifying animal species in camera trap images with high accuracy.
An early attempt at recognizing periodic movements in normal gray-scale image sequences can be found in \cite{Polana1997ijcv}.

Long-term observation using computer vision is one of the main challenges in wildlife monitoring, especially using camera traps, which operate on a limited power and storage budget. 
To save power, images are acquired sparingly. 
In contrast, event cameras offer the potential for low-power data acquisition.
We exploit data recorded in such a scenario and analyze oscillatory animal behavior.
Our proposed Fourier-based methods offer a novel approach to event camera-based action recognition, focusing on frequency patterns associated with specific actions to recognize wildlife behavior (in our case, of chinstrap penguins).
We envision that simple classifiers like the ones here proposed have the potential to enable online computer vision applications during long-term observation.

\subsubsection{Event Cameras and Oscillatory Motion in Nature}
\label{sec:related_work:evcamosc}
Oscillatory and repetitive patterns are widely present in nature and can be observed frequently as part of a variety of animal behaviors. 
A few examples include woodpeckers drilling with their bills for foraging and nest building \cite{copeyon1990technique}, tool use by chimpanzees \cite{boesch1994nut} and crows \cite{auersperg2011flexibility}, or the mating displays of peacock spiders \cite{girard2011multi}, birds of paradise \cite{dinsmore1970courtship} and penguins \cite{brunton2010ecstatic}.

Existing research highlights the substantial interest and potential in the emerging field of observing oscillatory motion with event cameras. 
For example, \cite{Pohle23visapp} tried to use the FFT to classify insect species based on wing-beat frequencies.
However, they found that the variability in wing-beat frequencies was very high, making this approach unsuitable for the intended classification task.
Concurrent unpublished work \cite{Takatsuka23biorxiv} utilizes FFT-based features for the classification of plankton in laboratory and lake environments, revealing that certain plankton species swim at particular frequencies.
Our work demonstrates a higher comprehensiveness level by carrying out experiments on four hours of annotated event data, thus ensuring the robustness of the proposed methods.

\section{Action Recognition with an Event Camera and the Fourier Transform}
\label{sec:method}

Many natural phenomena present an oscillatory character. 
When observed with cameras, such oscillations are recorded in the form of intensity changes on a pixel array. 
Specifically, the pixels of an event camera respond to intensity changes asynchronously, producing a stream of ``events'' at the same rate as the scene dynamics \cite{Posch14ieee,Gallego20pami}. 
Hence, event cameras are naturally fit to observe oscillatory phenomena (see examples \cite{Pfrommer22arxiv,Niwa23cvprw,Shiba23pami}). 

In the natural phenomenon under study, the penguin's ED consists of wing-flapping at an approximate frequency of two times per second.
We hypothesize that we can recognize this distinctive animal behavior by analyzing the data it produces when acquired by an event camera (\cref{fig:intro:eyecatcher}).
The key idea is that if the penguin flaps its wings, a sinusoidal signal appears on the event rate (number of intensity changes produced per second, for each polarity).
Naturally, periodic signals can be analyzed in the Fourier domain, which provides information about the frequency components of the signals. 
Hence, it is intuitive that it should be possible to recognize an ED simply by assessing the energy content of the signal in a band around the expected wing-flapping frequency (2~Hz). 

\subsection{Preprocessing}
\label{sec:method:preprocessing}
Formally, let $\cE = \{ e_i \}$ be the event data in a time interval of duration $d$.
An event $e_i = (x_i, y_i, t_i, \pol_i)$ contains the pixel coordinates $(x_i,y_i)$, timestamp $t_i$ and sign $\pol_i=\{+1,-1\}$ (i.e., polarity) of an intensity change of preset size $C$ (typically, 15\% relative change) \cite{Gallego20pami}.
The discrete event rate is a one-dimensional signal counting the number of events per time step $\Delta t > 0$. 
Splitting by polarity, the rates corresponding to the positive (ON) and negative (OFF) events are $r_{\text{ON}}[k]$ and $r_{\text{OFF}}[k]$, respectively, where $k$ denotes the sample number along the time axis.
Please note that a timestep $k$ is different from single events $e_i$, as mostly a high number of events happen within one time bin.
The rates by polarity $r_{\text{ON}}[k]$ and $r_{\text{OFF}}[k]$ 
are further combined into a signed (or ```polarity-aware'') event rate: 
\begin{equation}
\label{eq:event_rate}
r[k] \doteq r_{\text{ON}}[k] - r_{\text{OFF}}[k].
\end{equation}

The Fourier transform of the event rate is (\cref{fig:method:method})
\begin{equation}
R[f] \doteq \mathcal{F}\{r[k]\}.
\label{eq:rateFT}
\end{equation}
Next, we present ED classification criteria based on the characteristics of \eqref{eq:rateFT}.

\subsection{Energy-band Classifier}
\label{sec:method:energy_band}
Letting the lower and upper boundaries of a frequency band be $\flower$ and $\fupper$, respectively, 
the energy of $r[k]$ contained in this frequency band is
\begin{equation}
E_{\flower, \fupper} \doteq 2 \int_{\flower}^{\fupper} |R[f]|^2 df.
\label{eq:method:band_energy}
\end{equation}
In practice, we approximate the integral in \eqref{eq:method:band_energy} by the sum of discrete values of the FFT (also called $R$ by abuse of notation), $2 \sum_{f\in [\flower,\fupper]} |R[f]|^2$.

The event rate is susceptible to changes of environmental conditions, like the amount of noise for different lighting conditions.
Therefore, we normalize the band energy $E_{\flower, \fupper}$ with respect to the energy of the whole signal $E_{0,\infty}$ as:
\begin{equation}
    \hat{E}_{\flower, \fupper} \doteq E_{\flower, \fupper} \,/\, E_{0,\infty}.
    \label{eq:method:band_energy:normalized}
\end{equation}

Our first classifier is inspired by the Neyman-Pearson detector \cite{Neyman1933trs}. 
We propose using the normalized band energy \eqref{eq:method:band_energy:normalized} as the feature for classification based on a decision threshold.
If $\hat{E}_{\flower, \fupper} > \lambda$, we conclude that an ED is present.
We use $d$ seconds of event data from a penguin nest (region of interest, ROI) as observation time (in the experiments, $d=5$s).
The band parameters $\flower$, $\fupper$ and threshold $\lambda$ need to be tuned for the data at hand.
Details on parameter tuning can be found in \cref{sec:exp:implementational_details}.

\subsection{Full-spectrum Classifiers} 
\label{sec:method:spectrumclassifiers}
\begin{figure}[t]
    \centering
\begin{subfigure}{0.48\linewidth}
  \centering
  \includegraphics[trim={11.5cm 4.5cm 12cm 4.5cm},clip,width=0.8\linewidth]{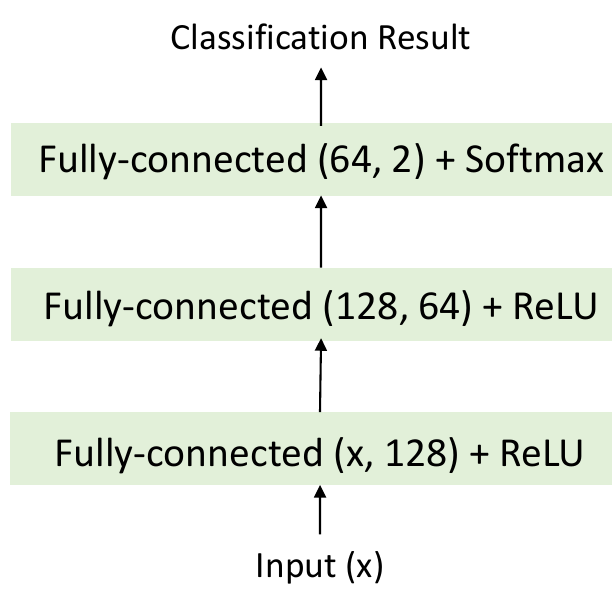}
  \caption{fc}
\end{subfigure}
\begin{subfigure}{0.48\linewidth}
  \centering
  \includegraphics[trim={11cm 4.8cm 11.5cm 1.8cm},clip,width=0.8\linewidth]{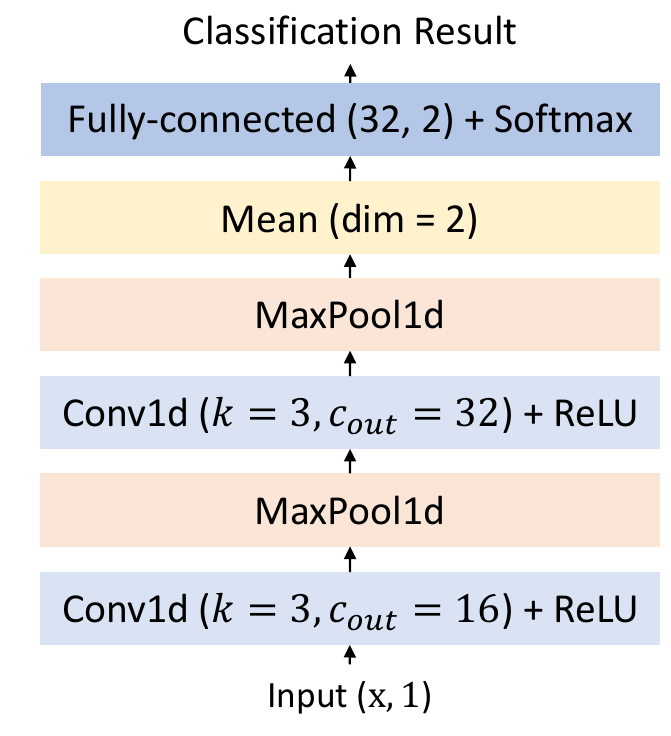}
  \caption{conv1d}
\end{subfigure}%
\caption{The architectures of the classification networks. 
Here, $x$ refers to the size of input vector.}
    \label{fig:app:network_architectures}
\end{figure}

In addition to the described intuitively designed classifier, we present two learning-based approaches (i.e., artificial neural networks ANNs). 
One classifier is based on a fully-connected layer and another one is based on two convolutional layers.
The input to both classifiers is the 1D array $|R(f)|$ containing the spectrum of the event rate (magnitude of the FFT), normalized by its peak value.
We refer to these two methods as ``FFT + fc'' and ``FFT + conv1d'', respectively.
The details of the network architectures are provided in \cref{fig:app:network_architectures}. 
The architecture and number of parameters for each model depend on the size of the 1D input signal (FFT spectrum or rate).
These architectures will also be used in the comparison methods of \cref{sec:exp:comparison_methods}.

\section{Dataset and Evaluation Metrics}
\label{sec:data}

\subsection{Dataset}
\begin{figure}[t]
   \centering
   \includegraphics[width=0.48\linewidth]{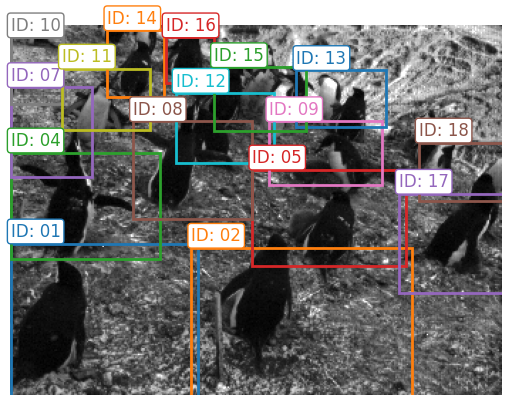}
   \gframe{\includegraphics[width=0.48\linewidth]{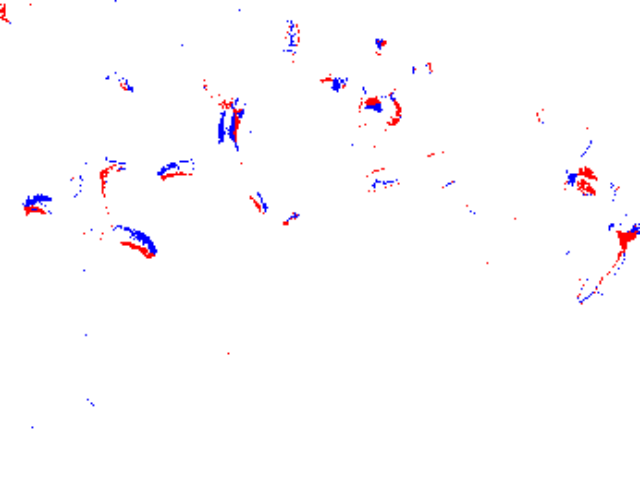}}
\caption{\emph{Visualization of the annotated regions of interest (ROIs, i.e., penguin nests)}.
Left: annotated bounding boxes around the individual penguin nests.
Right: events colored in red and blue (according to polarity) over a white canvas.
Data from \cite{Hamann24cvpr}.}
\label{fig:data:bboxes}
\end{figure}

We use a dataset of breeding chinstrap penguins recorded with a DAVIS346 event camera (\cref{fig:data:bboxes}).
The dataset in \cite{Hamann24cvpr} is annotated for the task of temporal action detection (TAD) \cite{Zhao20ijcv}.
Hence, the annotations consist of intervals of start and end times of the ED behavior in each nest (such regions of interest are pre-annotated with bounding boxes). 
The task we address is different: it is not a regression task (determining start and end times), but rather a classification one, on windows of duration $d$.
Therefore, to use the annotations from the ``Event Penguins'' dataset \cite{Hamann24cvpr}, we need to first compute the ground truth for our task, as follows.

\def\numSamples{N_\text{tot}}
\def\numPosSamples{N_\text{target}}
\def\truepos{\text{TP}}
\def\falsepos{\text{FP}}
\def\falseneg{\text{FN}}
We sample one $\langle$timestamp, label$\rangle$ pair every 33ms for each penguin nest.
The label is positive if the timestamp lies within an interval of an ED, and negative otherwise.
Our task is evaluated on samples of these pairs.
The corresponding input given a timestamp $t$ is the events of one nest within the interval $[t - d/2, t + d/2]$.
The action class (ED) only happens during short times, and most time intervals in the dataset belong to the ``background'' (BG) class.
Thus, the above data conversion produces $\numSamples = 427997$ samples, of which only $\numPosSamples = 10364$ are positive 
(i.e., 2.42\% of the samples, which implies a large class imbalance).

We split the annotated data (24 ten-minute sequences) into training and test sets.
An 80\% / 20\% split leads to five sequences in the test set (see \cref{tab:data:overview}).
The dataset contains scenes from various illumination and weather conditions (\cref{fig:exp:hdr}),
such as night and rain sequences.
\def\figmethodwidth{.48\linewidth}

\begin{table}[t!]
\centering
\caption{\emph{Data sequences in the test set.}}%
\label{tab:data:overview}
\begin{tabular}{l*{2}{S[detect-weight,detect-mode]}rr} 
\toprule
\text{Date \& time} & \text{Night}   & \text{Snow}  & \text{\#ED}  & \text{\#BG} \\ 
\midrule 
 7th Jan, 2:00      & \cmark & \xmark & 241 & 17695 \\
 7th Jan, 5:00      & \xmark & \cmark & 1307 & 16517 \\
 7th Jan, 8:00      & \xmark & \xmark & 85 & 17883 \\
 12th Jan, 17:26    & \xmark & \xmark & 1086 & 16898 \\
 15th Jan, 13:58    & \xmark & \xmark & 1 & 17824 \\
\bottomrule
\end{tabular}
\end{table}

\def\figWidth{0.31\linewidth}
\begin{figure}[t]
	\centering
    {\scriptsize
    \setlength{\tabcolsep}{1pt}
	\begin{tabular}{
	>{\centering\arraybackslash}m{0.25cm} 
	>{\centering\arraybackslash}m{\figWidth} 
    >{\centering\arraybackslash}m{\figWidth} 
	>{\centering\arraybackslash}m{\figWidth}}
 
        \rotatebox{90}{\makecell{Grayscale images}}
		&{\includegraphics[width=\linewidth]{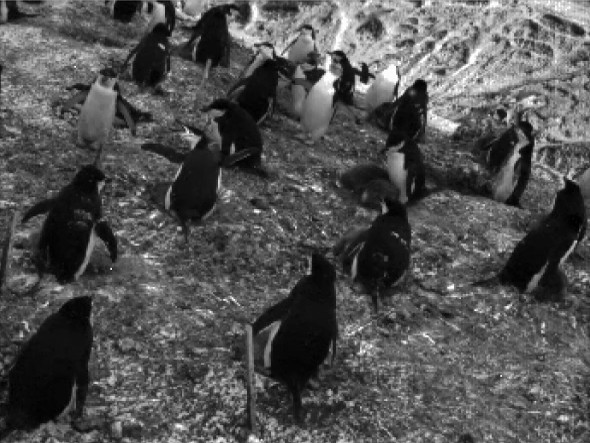}}
        &{\includegraphics[width=\linewidth]{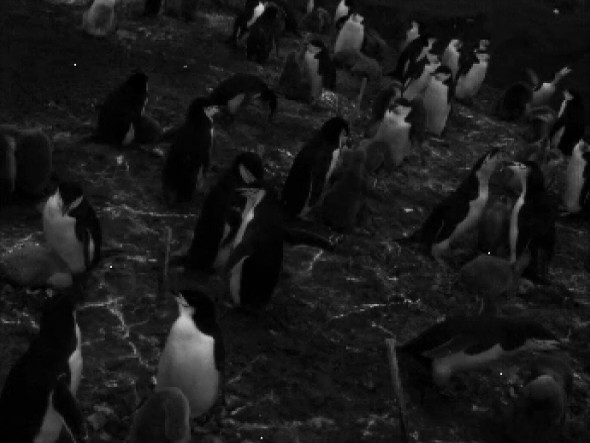}}
        &{\includegraphics[width=\linewidth]{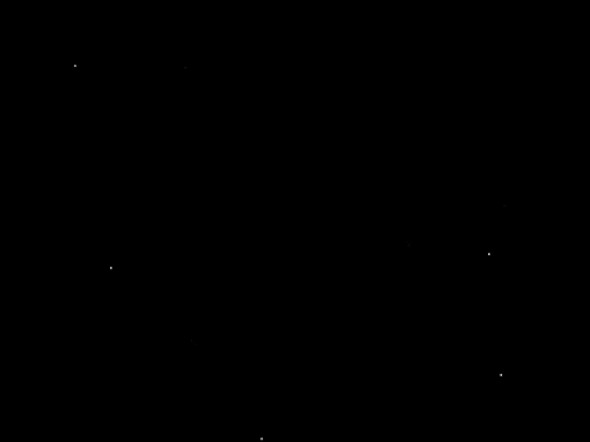}}\\
        
		\rotatebox{90}{\makecell{Events}}
        &\gframe{\includegraphics[width=\linewidth]{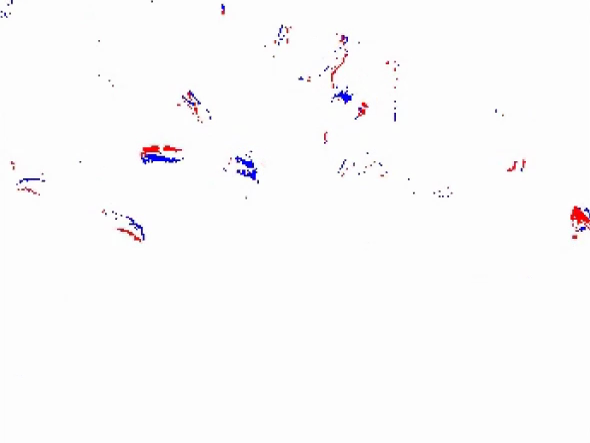}}
        &\gframe{\includegraphics[width=\linewidth]{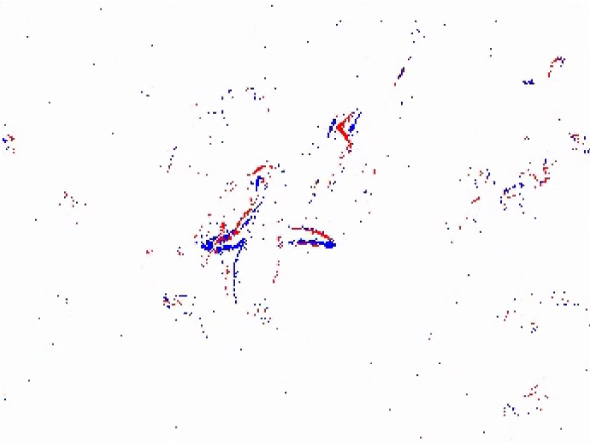}}
        &\gframe{\includegraphics[width=\linewidth]{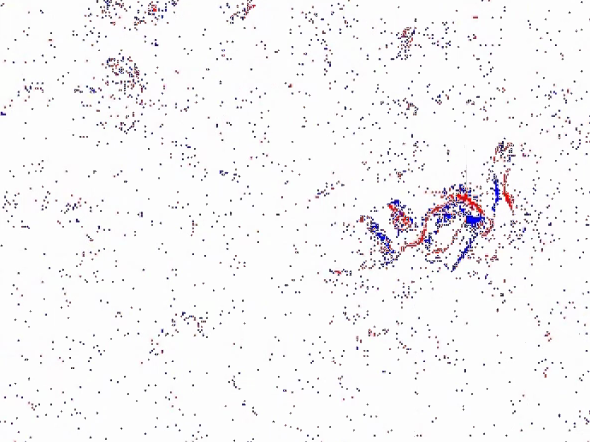}}\\
  
		& \textbf{(a)} Precipitation (snow)
        & \textbf{(b)} Dim light
		& \textbf{(c)} Night
	\end{tabular}
	}
	\caption{Examples of EDs recorded by events and grayscale images for different illumination conditions \cite{Hamann24cvpr}.
    The high dynamic range of events is a clear advantage in this application.}
	\label{fig:exp:hdr}
\end{figure}

\subsection{Evaluation Metrics}
\label{sec:FOne_random_classifier}
Basic evaluation metrics of a binary task like ED classification are \emph{precision} and \emph{recall}.
Precision measures the proportion of correctly predicted positive observations out of all positive predictions made by the model. 
It is given by the formula: $\text{Precision} = \truepos{} / (\truepos{} + \falsepos{})$,
where $\truepos{}$ and $\falsepos{}$ stand for True Positive and False Positives, respectively.
Recall measures the proportion of correctly predicted positive observations out of all actual positive observations. It is given by:
$\text{Recall}= \truepos{} / (\truepos{} + \falseneg{})$,
where $\falseneg{}$ means False Negative.

The \emph{F1 score} combines precision and recall via their harmonic mean:
F1 Score = 2 ((Precision $\cdot$ Recall) / (Precision + Recall)).
We use this blended metric in the experiments, as it can already take into account the large class imbalance.

\subsubsection*{F1 score of a random classifier}
Assuming a random classifier, i.e., flipping a fair coin for each of the $\numSamples$ samples,
gives $\truepos{} = \falseneg{} = \numPosSamples / 2$ and $\falsepos{} = (\numSamples - \numPosSamples)/2$.
From here, we obtain $\text{Recall}=0.5$
and $\text{Precision}= \truepos{} / (\truepos{} + \falsepos{}) 
= (\numPosSamples / 2) / ((\numPosSamples / 2) + ((\numSamples - \numPosSamples)/2)) 
= \numPosSamples / \numSamples = 0.0242$, 
which is considerably smaller than recall due to the large class imbalance. 
The F1 score is the harmonic mean of 0.5 and 0.0242, which is 0.046, as it is dominated by the smallest number between precision and recall.
Hence, the F1 score of a random classifier on this dataset is close to zero.

\section{Experiments}
\label{sec:exp}

\subsection{Implementation Details}
\label{sec:exp:implementational_details}

The event rate is obtained with a bin width of $0.01s$, which is chosen small enough to resolve the highest frequency of the ED that we aim to observe (a sensitivity analysis is given in \cref{sec:exp:sensitivity}).
To obtain modest results, the three parameters ($\flower, \fupper, \lambda$) of the energy-band classifier (\cref{sec:method:energy_band}) method need to be tuned.
To tune the parameters we search within reasonable bounds of the parameter space (guided by domain knowledge) 
by randomly sampling triplets of ($\fmid, b, \lambda$), where $\fmid$ is the middle frequency and $b$ is the bandwidth in \eqref{eq:method:band_energy}.
The lower and upper bounds can be calculated by $\flower = \fmid - b/2$ and $\fupper = \fmid + b/2$.
Random samples are drawn from the intervals $\fmid \in [1.8, 6]$, $b \in [0.3, 1.8]$ and $\lambda \in [0.05, 0.3]$.
Furthermore, we consider two cases: one where a common set of parameters is determined for all ROIs, and another one where each ROI has its own set of parameters.
Recall that the feature used for classification is the normalized band energy \eqref{eq:method:band_energy:normalized}. 
We choose the parameters that lead to the best results on the training set and report the results on the test set.

The ANN-based classifiers are trained using an Adam optimizer \cite{Kingma15iclr} with a learning rate of 0.003, a weighted cross-entropy loss, and a batch size of 32.
The experiments were performed on a computer with an Nvidia Tesla V100S GPU and an Intel Xeon 4215R CPU.

\subsection{Baseline Methods}
\label{sec:exp:comparison_methods}

\textbf{Overview.} For comparison we provide \emph{four additional} methods, 
with the first three also based on the proposed idea of summarizing the data into a low-dimensional signal. 
While these are, to some extent novel in the context of event cameras, this paper focuses on the distinctive contribution of combining event cameras with Fourier analysis, 
and therefore we refer to these four additional methods as baselines.
The methods are explained next, sorted by their computational complexity.
All comparison methods use the same number of events, like the FFT-based methods ($d$ = 5s).
If not specifically mentioned, training details are as described in \cref{sec:exp:implementational_details}.

\textbf{Energy-based classifier}. 
The first method is a simple heuristic that we term energy-based classifier. 
It determines the normalized energy of the events (effectively counting the events in the window) and assumes an ED is present if the energy is above a set threshold.
It is similar to the energy-band classifier in \cref{sec:method} but uses the whole signal energy $E_{0,\infty}$ as a feature for classification.
The Neyman-Pearson classification threshold is found in a similar approach as for the energy-band classifier, randomly sampling candidate thresholds from a reasonable interval.
The threshold with the best result on the training data is used for model evaluation.

\textbf{Rate-based classifiers}.
The following comparison methods are learning-based.
These second and third baseline methods use similar fully connected (fc) and convolutional (conv1d) ANNs as those in \cref{sec:method}; the difference lies in the pre-processing step.
We test several models using pre-processed 1D representations (i.e., vectors) as input.
Specifically, we test two architectures (dense and convolutional) and two event representations (rate and Fourier-spectrum of the rate).
This results in a total of four combinations:
the models that use the event rate as input representation (1D), which are termed ``rate + fc'' and ``rate + conv1d'',
and the models that use the Fourier transformer signal as input, which are termed ``FFT + fc'' and ``FFT + conv1d''.
In the latter, the input to the ANNs is the signed event rate $r[k]$, as opposed to the Fourier spectrum.
The only architectural difference between the ANNs is the number of input nodes of the first fully connected layer.

\textbf{2D CNN classifier}. 
Lastly, we compare against a learning-based residual network ResNet18 \cite{He16cvpr} that is trained to recognize ED from event histograms \cite{Maqueda18cvpr} (image-like representations of event data), and therefore it is termed ``2D CNN''.
Data augmentation is performed during training by random horizontal flipping of the samples.
We use a Stochastic Gradient Descent (SGD) optimizer with a learning rate of 0.002 and a momentum of 0.9 at a batch size of 128.
The ResNet was pre-trained on the ImageNet dataset \cite{Deng09cvpr}.

\emph{Remark}: Note the different space-time aggregation character of the methods: 
the ``2D CNN'' (ResNet18) method aggregates all temporal information of the events into an image (by pixel-wise event count during the observation interval $d$), while all rate- and spectrum-based methods aggregate all spatial information within the ROIs. 
Finally, the energy-based classifier aggregates all spatial and temporal information into a single number.

\subsection{Comparison with the State of the Art}
\label{sec:exp:results}
\begin{table}[t!]
\centering
\caption{\emph{Comparison of all methods on all ROIs.}
}
\adjustbox{max width=\linewidth}
{%
\setlength{\tabcolsep}{4pt}
\begin{tabular}{lr*{3}{S[detect-weight,detect-mode]}} 
\toprule
Method   &  \text{\#Params}  & \text{Precision}  & \text{Recall}  & \text{F1} \\ 
\midrule 

Random classifier  & 1 & 0.024 & 0.5 & 0.046  \\ 

Energy classifier  & 1 & 0.09926 & 0.1326 & 0.1135  \\

Energy-band classifier 1 (Ours) & 3  & 0.4328  &  0.3508 & 0.3875  \\

Energy-band classifier 18 (Ours) &  54    & 0.5221   &   0.5655     &  0.543 \\

FFT + fc (Ours) & 40600  & 0.452 &	0.7408 &	0.5614 \\ 

FFT + conv1D (Ours) & 1700  & 0.3903 & 0.6467 & 0.4868 \\ 

rate + fc   & 72500  & 0.3875 & 0.715 & 0.5026 \\ 

rate + conv1D   & 1700  & 0.462 & 0.7728 & 0.5783 \\ 

2D CNN (ResNet18) & \text{11.4M}  & 0.79973    &   0.67279     &  0.7227 \\
\bottomrule
\end{tabular}
}
\label{tab:exp:compare_energy_fft_cnn}
\end{table}

\def\figmethodwidth{.48\linewidth}
\begin{figure}[t]
   \centering
   \includegraphics[trim={10px 0 40px 35px},clip,width=0.9\linewidth]{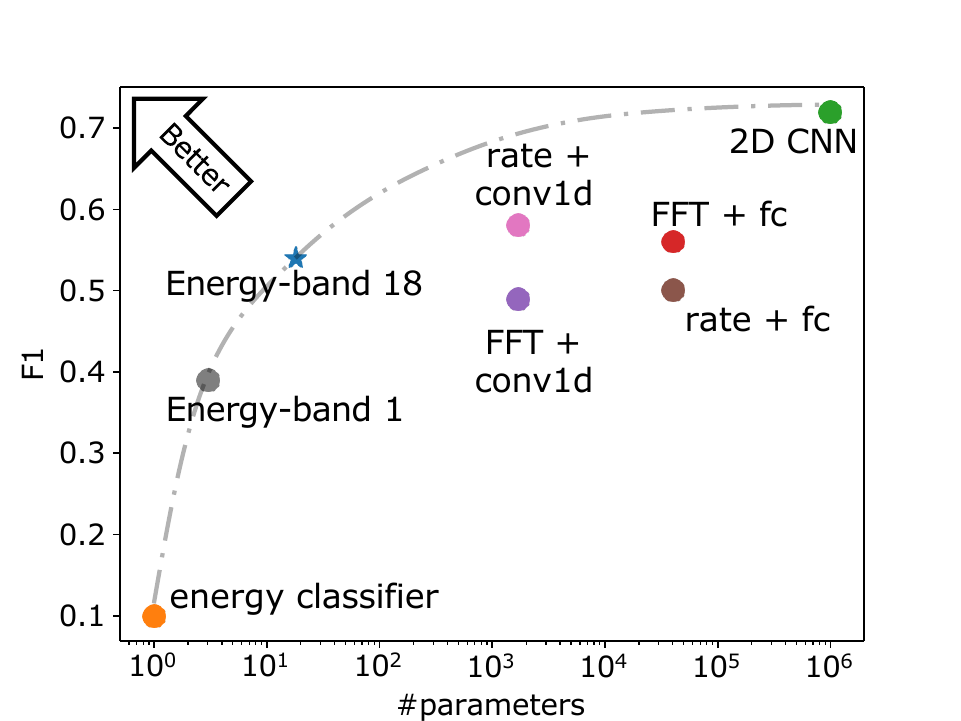}
\caption{Visualization of the F1 score and number of parameters of the compared methods (feature extractor and classifier).}
\label{fig:exp:f1_vs_params}
\end{figure}

\begin{figure*}
    \centering
    \includegraphics[trim={2.5cm 6.7cm 2.5cm 6cm},clip, width=\linewidth]{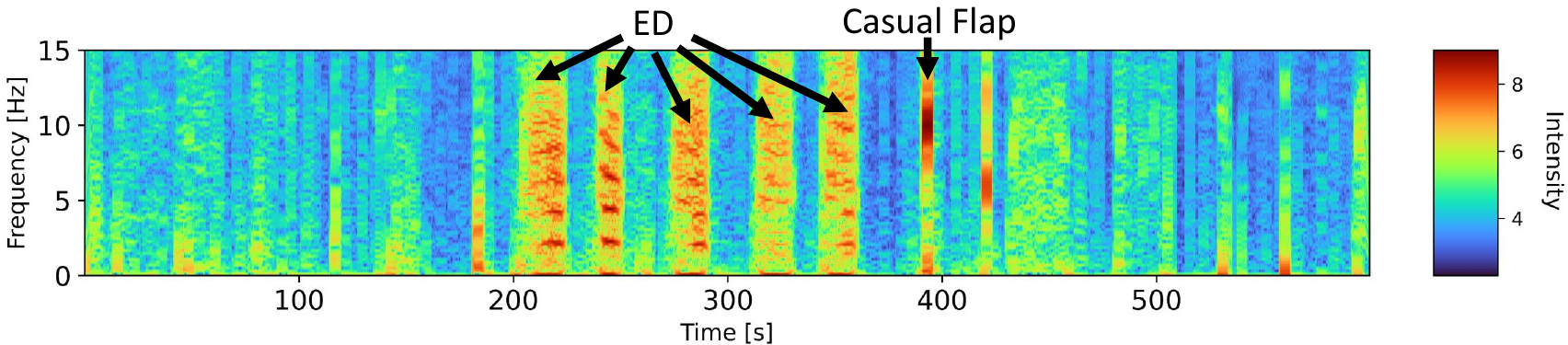}
    \caption{\emph{Spectrogram}.
    The heat map represents the frequency content of the signed event rate of one ROI (penguin nest) over a 10-minute interval.
    The ecstatic display (ED) shows a clear and characteristic pattern (e.g., energy distribution in frequency bands), 
    while other wing flaps (casual flap), which are not ED, have different patterns.}
    \label{fig:exp:spectrogram}
\end{figure*}

\Cref{tab:exp:compare_energy_fft_cnn} shows a comparison between all above-mentioned methods: 
the FFT-based and the baseline methods. 
The performance comparison is further visualized in \cref{fig:exp:f1_vs_params}.
The \mbox{1-parameter} classifier fails: simply measuring the energy of a time window of events does not yield good ED classification results (F1 score = 0.114, which is 2.467 times larger than that of the random classifier, but it is still small).
On the other end, the largest capacity model (2D CNN), with 11.4 million learned parameters, achieves the highest F1 score (0.72).
In between these extremes, we find the rate-based and FFT-based classifiers.

The next best model to the energy-based 1-parameter classifier is the 3-parameter energy-band classifier (with all ROIs sharing the same parameters, called ``Energy-band 1''). 
There is a considerable performance increase, from 0.11 to 0.39 in the F1 score, but having a single set of parameters for all ROIs is suboptimal, as each penguin nest is different from the others (concerning the camera distance, appearance, penguin motion, etc.).
The energy-band classifier with different parameters per ROI (called ``Energy-band 18'') yields a further performance increase (0.54 in F1 score). 
Comparing this model against the other two models with a similar F1 score of about 0.55, i.e., ``FFT + fc'' and baseline method ``rate + conv1d'', 
we observe that the energy-band classifier has only 54 parameters versus the 1700 and 40600 parameters of the other two models. 
Hence, it is at least 35 times more efficient in number of learned parameters (i.e., model size) than competing methods, 
and, it is also more interpretable because the 3 parameters per ROI have a clear physical meaning. 
Compared to the F1 score of the random classifier (0.046, \cref{sec:FOne_random_classifier}), an F1 score of 0.54 is significantly larger.
Likewise, while the performance is inferior to the 2D CNN one, the difference in number of parameters is mesmerizing: 54 vs.~11.4 million, i.e., five orders of magnitude smaller.

Building on the interpretability advantage of the energy-band classifiers, 
their obtained middle frequencies lie around 2.2~Hz, which indicates that the detector reacts to the expected frequency of the ecstatic display.
The results of the energy-based classifier show that the model assumption is reasonable, the signal energy caused by the ecstatic display is measurably higher than the rest of the signal.
The results show the robustness against responses in the Fourier transform caused by other phenomena.

\subsection{Qualitative Analysis}
\begin{figure}[t]
	\centering
    {
    \setlength{\tabcolsep}{2pt}
	\begin{tabular}{
	>{\centering\arraybackslash}m{0.3cm} 
    >{\centering\arraybackslash}m{0.17\linewidth} 
	>{\centering\arraybackslash}m{0.37\linewidth} 
	>{\centering\arraybackslash}m{0.37\linewidth}}
 
        \rotatebox{90}{\makecell{Ecstatic Display}}
		&{\includegraphics[width=\linewidth]{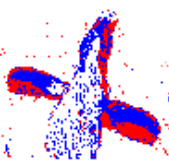}}
        &{\includegraphics[width=\linewidth]{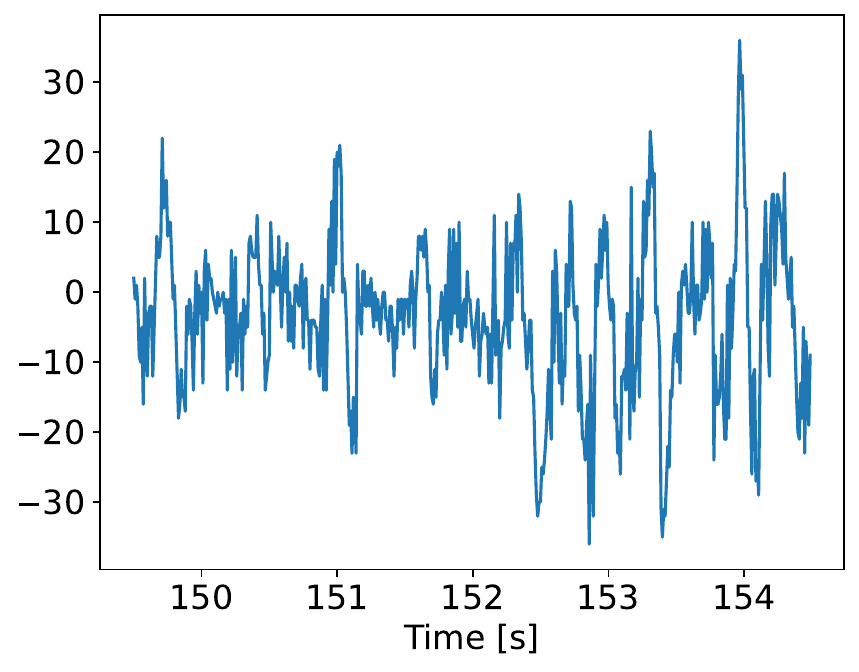}}
        &{\includegraphics[width=\linewidth]{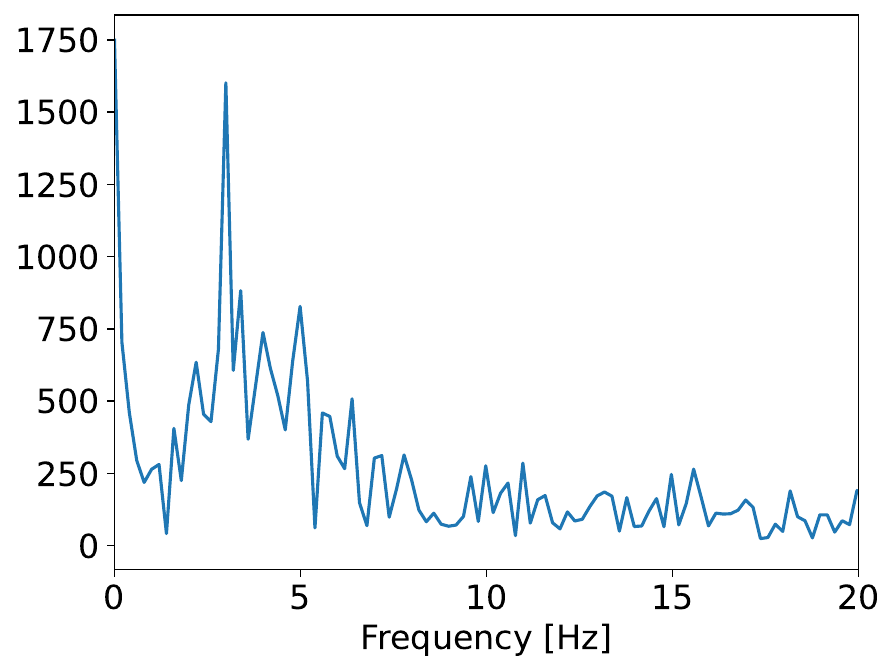}}
        \\
        \rotatebox{90}{\makecell{Casual flap}}
		&{\includegraphics[width=\linewidth]{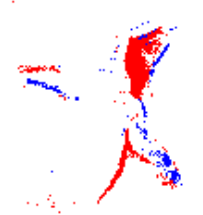}}
		&{\includegraphics[width=\linewidth]{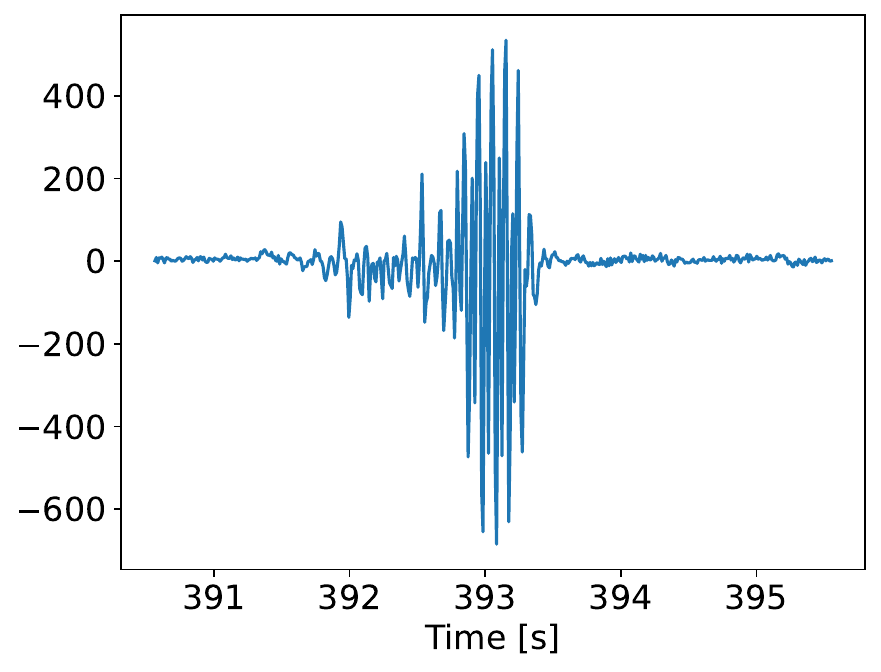}}
        &{\includegraphics[width=\linewidth]{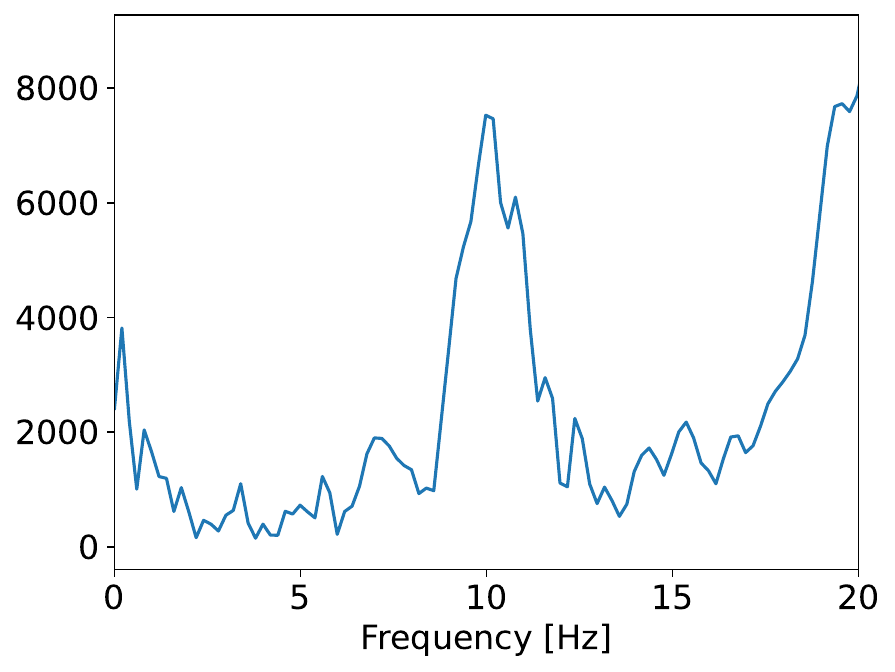}}
        \\
        \rotatebox{90}{\makecell{Head shake}}
		&{\includegraphics[width=\linewidth]{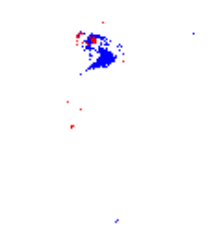}}
		&{\includegraphics[width=\linewidth]{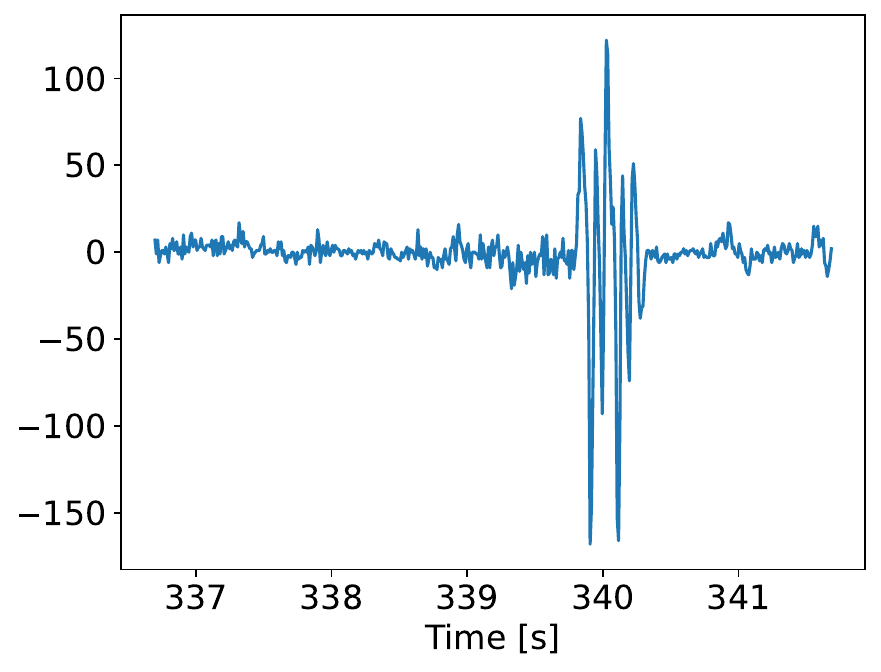}}
        &{\includegraphics[width=\linewidth]{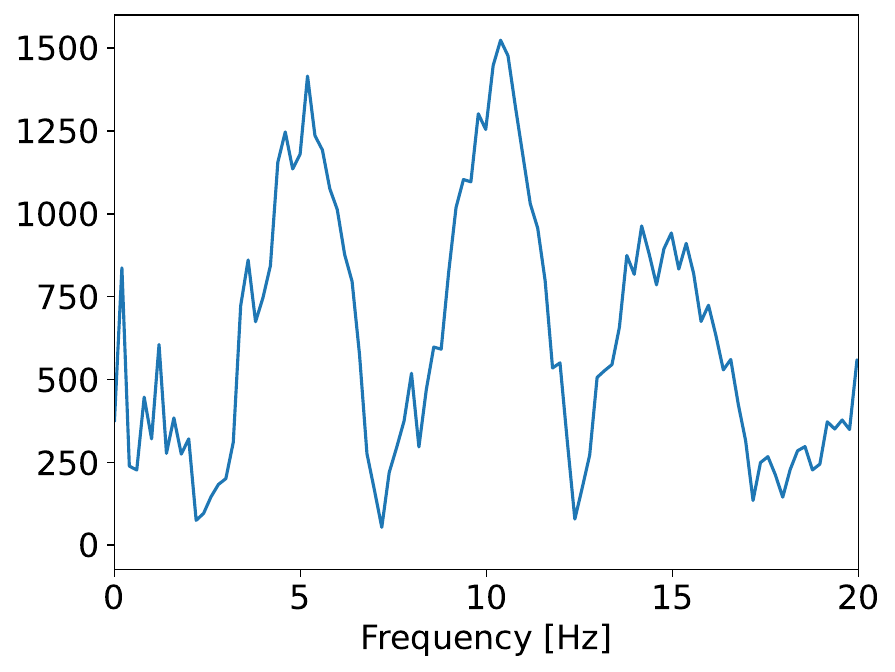}}
        \\
        \rotatebox{90}{\makecell{Background}}
		&{\includegraphics[width=\linewidth]{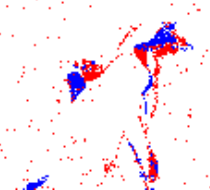}}
		&{\includegraphics[width=\linewidth]{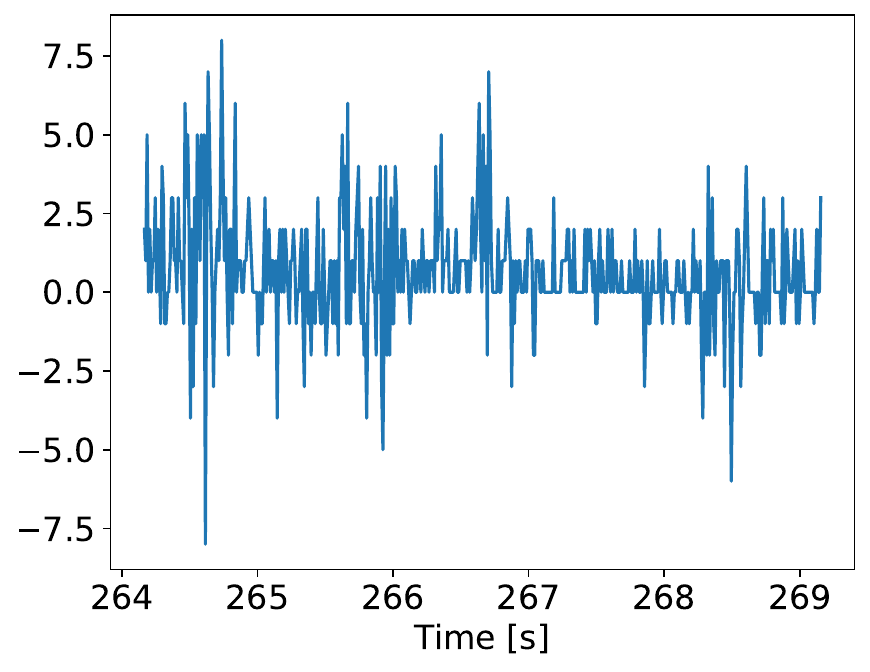}}
        &{\includegraphics[width=\linewidth]{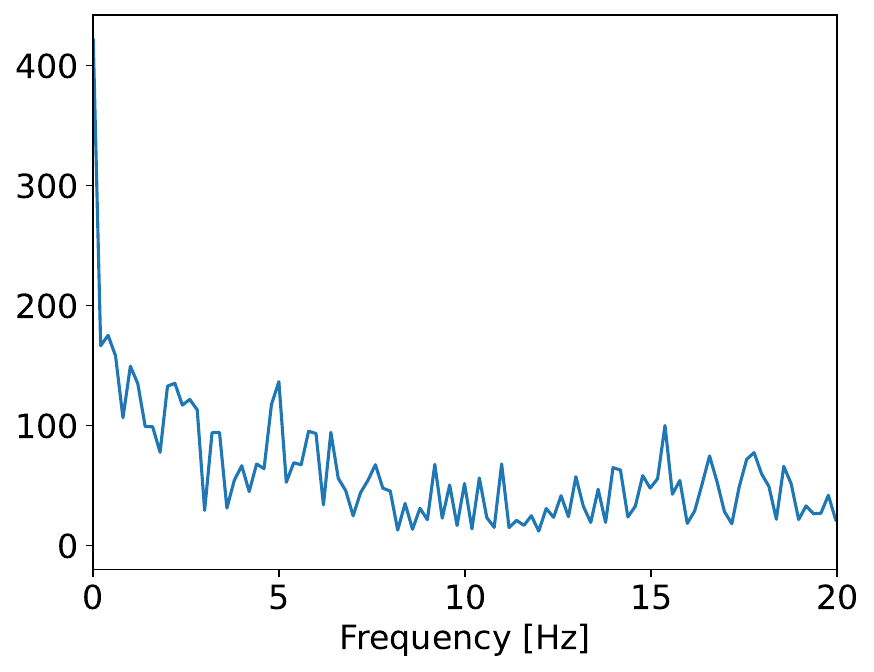}}
        \\

		& \textbf{(a)} Events
		& \textbf{(b)} Signed event rate
        & \textbf{(c)} Spectrum
	\end{tabular}
	}
	\caption{Examples of different penguin behaviors (rows) and different representations of the signal (columns).
    Note the large differences in vertical range of the plots across rows, specially in the spectrum plots.
    }
	\label{fig:exp:qualitativepenguinbehaviors}
\end{figure}

\Cref{fig:exp:spectrogram} gives a visual impression of how and why the proposed method works. 
It shows a spectrogram of the signed event rate of one nest over ten minutes.
During this time the penguin performs five ecstatic displays.
The ED behavior shows a unique pattern, which is distinguishable from other similar behaviors, such as a casual wing flap (marked in \cref{fig:exp:spectrogram}).
The latter happens at a higher frequency, which aligns with the pattern visible in the spectrogram.
This visualization delivers an intuition why classification with simple criteria in the Fourier domain is possible and how the nature of the event data inherently helps to solve the problem despite its limitations (\cref{sec:limitations}). 

Furthermore, \cref{fig:exp:qualitativepenguinbehaviors} shows examples of different behaviors, and their respective event rates and spectra.
Like the spectrogram, the visualization provides intuitive explanations of why the Fourier-based methods provide meaningful pre-processing.
The spectrum of the ED shows a clear peak around 2~Hz, while other behaviors, like the casual flap or the head shake, have the energy spectral density distributed over higher frequencies.
For completeness, an example of background pattern is also shown in \cref{fig:exp:qualitativepenguinbehaviors}, whose spectrum spreads more uniformly over all frequencies and has considerably smaller energy.

\subsection{Sensitivity and Ablation Studies}
\label{sec:exp:sensitivity}

\begin{table}[t!]
\centering
\caption{\emph{Sensitivity} of precision, recall and F1 score 
with the bin width and window duration (in seconds) for the Energy-band classifier 1.}
\adjustbox{max width=\columnwidth}{%
\begin{tabular}{l*{4}{S[detect-weight,detect-mode]}} 
\toprule
\text{Bin width} & \text{Window duration}      & \text{Precision}  & \text{Recall}  & \text{F1} \\ 
\midrule 
 0.01    &   5    & 0.5662  & 0.5216  & 0.543  \\
 0.005   &   5    & 0.6449  & 0.4124  & 0.5031 \\
 0.1     &   5    & 0.5677  & 0.4921  & 0.5272 \\
 0.01    &   1    & 0.4475  & 0.4626  & 0.4549 \\
 0.01    &   10   & 0.5313  & 0.5218  & 0.5265 \\
\bottomrule
\end{tabular}
}
\label{tab:exp:sensitivity}
\end{table}

\begin{table*}[t]
\centering
\caption{Results of the energy-band classifier for each of the 18 penguin nests considered in the dataset.}
\adjustbox{max width=\textwidth}{%
\setlength{\tabcolsep}{4pt}
\begin{tabular}{l*{18}{S}}

\toprule
  Metric                & \text{N01}  & \text{N02}  & \text{N03} & \text{N04} & \text{N05} & \text{N06} & \text{N07} & \text{N08} & \text{N09} & \text{N10} & \text{N11} & \text{N12} & \text{N13} & \text{N14} & \text{N15} & \text{N16} & \text{N17} & \text{N18} \\ 
\midrule 
\textbf{Precision} & 0.2986 & 0.9375 & 0.9036 & 0.7531 & 0.8480 & 0.2802 & 0.4615 & 0.5969 & 0.6905 & 0.5690 & 0.5105 & 0.3215 & 0.5946 & 0.4643 & 0.7399 & 0.3039 & 0.9221 & 0.4823 \\
\textbf{Recall} & 0.5291 & 0.7394 & 0.7282 & 0.5701 & 0.6709 & 0.6915 & 0.7000 & 0.6500 & 0.6905 & 0.6875 & 0.4569 & 0.6412 & 0.2785 & 0.2407 & 0.3710 & 0.2719 & 0.7978 & 0.7317 \\
\textbf{F1} & 0.3818 & 0.8268 & 0.8065 & 0.6491 & 0.7491 & 0.3988 & 0.5563 & 0.6223 & 0.6905 & 0.6226 & 0.4822 & 0.4283 & 0.3793 & 0.3171 & 0.4942 & 0.2870 & 0.8554 & 0.5814 \\
\bottomrule
\end{tabular}

}
\label{tab:exp:results_per_roi}
\end{table*}

Next, we further characterize the energy-band classifier. 

\Cref{tab:exp:sensitivity} presents the results of a sensitivity study conducted to analyze the influence of different parameters. 
In this study, we tune two parameters: the bin width $\tau_{\delta}$ and the window duration $d$. 
The bin width plays a role in determining the granularity of the event rate and affects the frequency bins. 
A comparison between two bin width values, 0.1 and 0.005 seconds, reveals slightly worse performance; however, the impact on the results is not substantial for the tested parameters.

The second parameter considered in this study is the window duration $d$. We present additional results for $d = 1$s and $d = 10$s. For $d = 1$s, the performance is 9\% worse, which seems reasonable given that the approximate frequency of the wing flap is 2 Hz. A window duration of 1 second may not capture about two periods of a wing flap, which appears insufficient. On the other hand, when $d = 10$s, the results are only slightly worse by 2\%. This finding suggests that increasing the window duration beyond $d = 5$s is not beneficial. 
This observation is consistent with the dataset statistics showing that the average duration of an ecstatic display is about 8 seconds \cite{Hamann24cvpr}. 

\Cref{fig:data:bboxes} shows the bounding boxes of the different nests.
It becomes apparent that the choice of the bounding box is an important factor, as it determines which events are taken into account for the classification decision.
\Cref{tab:exp:results_per_roi} confirms this intuition: classification results vary widely among different ROIs,
which are roughly indexed according to depth with respect to the camera (starting from the foreground and increasing towards nests further in the back).
\Cref{tab:exp:results_per_roi} reveals a general trend: nests far from the camera achieve worse classification scores.
Given the large dependence on the choice of ROI and the high overlap as seen in \cref{fig:data:bboxes}, we expect that better recordings (e.g., with decreasing visual overlap between the nests) will enable better classification results.
\Cref{tab:exp:ablation:rois} shows that indeed classification can be improved significantly if only the best ROIs are taken into account.
For the best two nests, the achieved F1 score is 82\%.

\subsection{Variations of the Estimation Method}

The method in \cref{sec:method:preprocessing,sec:method:energy_band} is just one way to estimate the band energy of an event rate signal.
However, our work is not limited to the explained specifics but allows for variations.
Let us present two adaptions of our method that achieve similar results as that in \cref{sec:method}.

\paragraph{Variation of the event rate signal}
One variation pertains to the calculation of the event rate.
We found that our method is sensitive to the mean (i.e., DC component) of the event rate signal.
\Cref{tab:exp:ablation} shows the results for the Energy-band 18 classifier with and without using polarity.
In the latter case it is calculated as $s[k] \doteq r_{\text{ON}}[k] + r_{\text{OFF}}[k]$.
The result demonstrates that this method yields an F1-score of only 0.08, indicating that it fails to provide accurate predictions.
Comparing both cases, we observe a significantly lower DC component in the event rate that considers the polarity.
Positive and negative events may partially cancel each other.
The results indicate that this effect helps the stability of the method.
However, we can achieve a similar effect by subtracting the mean $\mu$ of the rate 
$\tilde{s}[k] = s[k]-\mu(s[k])$.
\Cref{tab:exp:alterations} shows the results for using $\tilde{s}[k]$ (column ``Zero-mean rate'').
They confirm that using $\tilde{s}[k]$ provides a similar performance to the polarity-aware event rate $r[k]$.

\paragraph{Spectral Variation, Welch's method} 
At its core, the energy-band classifier relies on the estimation of the power spectral density (PSD).
The method described in \cref{sec:method}, which computes the square magnitude of the FFT of the signal, is also known as Periodogram.
It is a simple method and it provides sufficient estimates for our use-case.
Naturally, it is possible to use more sophisticated methods for the estimation of the PSD.
\Cref{tab:exp:alterations} provides results for using Welch's method \cite{Welch67tau}.
In comparison, both methods (1st and 3rd columns) are very similar, which proves that our method doesn't rely on the specific PSD estimation scheme.
On the tested data, the additional complexity of Welch's method does not pay off since it does not improve the results.

\begin{table}[t!]
\centering
\caption{\emph{Variation of the results of the enerby-band classifier with respect to the number of ROIs (i.e., nests) considered.}}
\adjustbox{max width=\columnwidth}{%
\begin{tabular}{l*{3}{S[detect-weight,detect-mode]}} 
\toprule
\text{\#ROIs} &  \text{Precision}  & \text{Recall}  & \text{F1} \\ 
\midrule 
2        &   0.9088   &  0.7438     & 0.8181  \\
4        &   0.876    &  0.712      & 0.7855  \\
8        &   0.7304   &  0.6728     & 0.7004  \\
12       &   0.6486   &  0.597      & 0.6217  \\
18 (all) &   0.5221   &   0.5655    &  0.543  \\
\bottomrule
\end{tabular}
}
\label{tab:exp:ablation:rois}
\end{table}

\begin{table}[t!]
\centering
\caption{\emph{Ablation studies}.  
Variation of precision, recall and F1 score with respect to event polarity.}
\adjustbox{max width=\columnwidth}{%
\begin{tabular}{l*{3}{S[detect-weight,detect-mode]}} 
\toprule
\text{Polarity}   & \text{Precision}  & \text{Recall}  & \text{F1} \\ 
\midrule 
\cmark    & 0.5221  & 0.5655  & 0.543   \\
\xmark    & 0.0507  &  0.1509 & 0.0759  \\
\bottomrule
\end{tabular}
}
\label{tab:exp:ablation}
\end{table}
\begin{table}[t!]
\centering
\caption{\emph{Variations} of the estimation method in \cref{sec:method}.}
\adjustbox{max width=\columnwidth}{%
\begin{tabular}{l*{3}{S[detect-weight,detect-mode]}} 
\toprule
& \multicolumn{3}{c}{F1} \\
\cmidrule(lr){2-4}
\text{Method}  & \text{Signed event rate (\cref{sec:method})}  & \text{Zero-mean rate}  & \text{Welch's} \\ 
\midrule 
Energy-band 1  &    0.3875                                       &    0.4361              &    0.3805     \\
Energy-band 18 &    0.5317                                       &    0.49626             &    0.5151     \\
FFT + conv1D   &    0.4868                                       &    0.402               &    0.4628     \\
FFT + fc       &    0.5614                                       &    0.5498              &    0.5519     \\
\bottomrule
\end{tabular}
}
\label{tab:exp:alterations}
\end{table}

\section{Limitations}
\label{sec:limitations}

The 1D rate signal that we have used as a summarizing ``feature'' for several methods discards the spatial information within the nest.
This implies that the energy-band classifier cannot distinguish two shakes of the same frequency at different positions (e.g., head and wing). 
Currently, it relies on the distinctiveness and separation of the actions in the chosen feature space to avoid such potential ambiguities.
Future work could investigate approaches that combine partial spatial information (akin to the 2D CNN method) with the used temporal information (signed event rate) while being succinct in model size.

Our approach also inherits some limitations from the dataset and the way the task is posed. 
Namely, the overlap of bounding boxes can introduce challenges in distinguishing multiple objects in the scene.
A simple solution would be to discard nests in the background (see \cref{tab:exp:ablation:rois}).
Likewise, recordings with a better viewpoint to single out individuals could lead to improved reliability.

\section{Discussion}
\label{sec:discussion}

The experiments on four hours of annotated event data demonstrate the robustness of the proposed methods.
The energy-band method utilizes simple features derived from the Fourier Transform, a highly relevant and efficient analysis tool.
By leveraging the FFT, we facilitate the feature extraction process within a physically interpretable domain, enabling the creation of a reduced dimensional model that preserves class separability.
The success of this method suggests that further exploration of alternative efficient harmonic analysis tools beyond the Fourier Transform (e.g., Wavelets), could yield additional advancements.

\begin{figure}[t]
	\centering
    {
    \setlength{\tabcolsep}{2pt}
	\begin{tabular}{
    >{\centering\arraybackslash}m{0.05\linewidth} 
	>{\centering\arraybackslash}m{0.45\linewidth} 
	>{\centering\arraybackslash}m{0.45\linewidth}}

        &{\includegraphics[width=.7\linewidth]{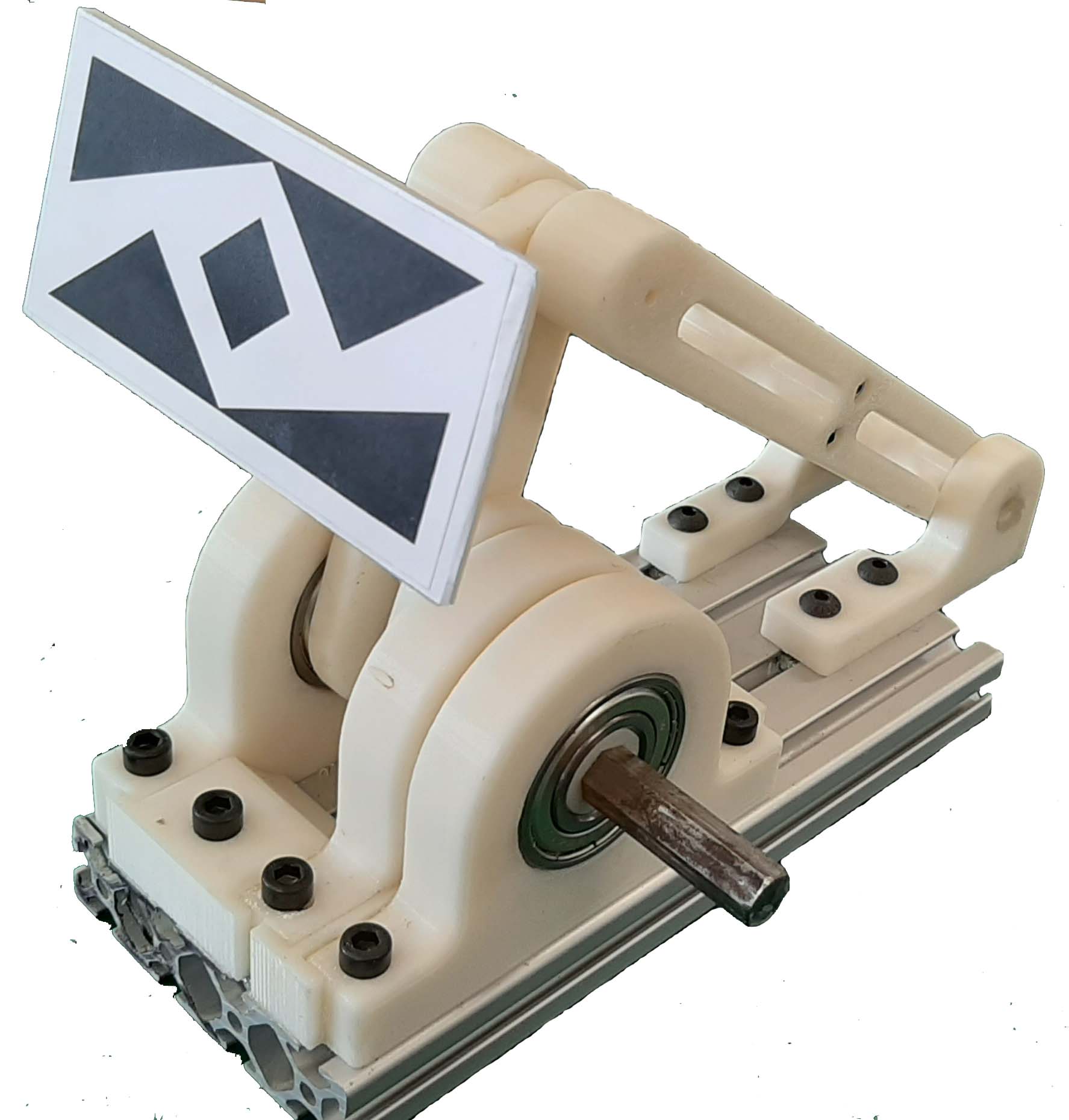}}
        &{\includegraphics[width=.9\linewidth]{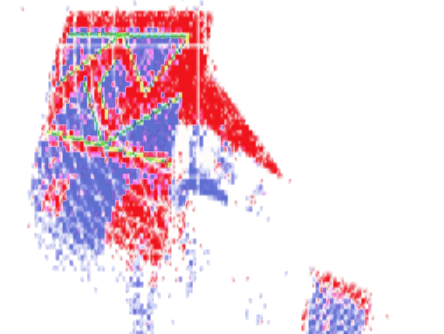}}
        \\
        & (a) Motor setup \cite{Chamorro20bmvc} & (b) Example events \cite{Chamorro20bmvc}
        \\[2ex]

        \rotatebox{90}{\makecell{10.9 Hz}}
		&{\includegraphics[width=\linewidth]{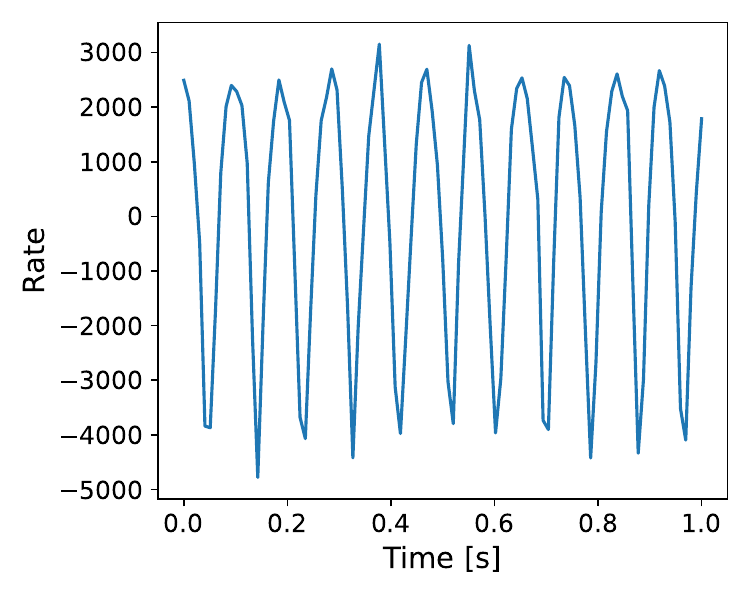}}
        &{\includegraphics[width=\linewidth]{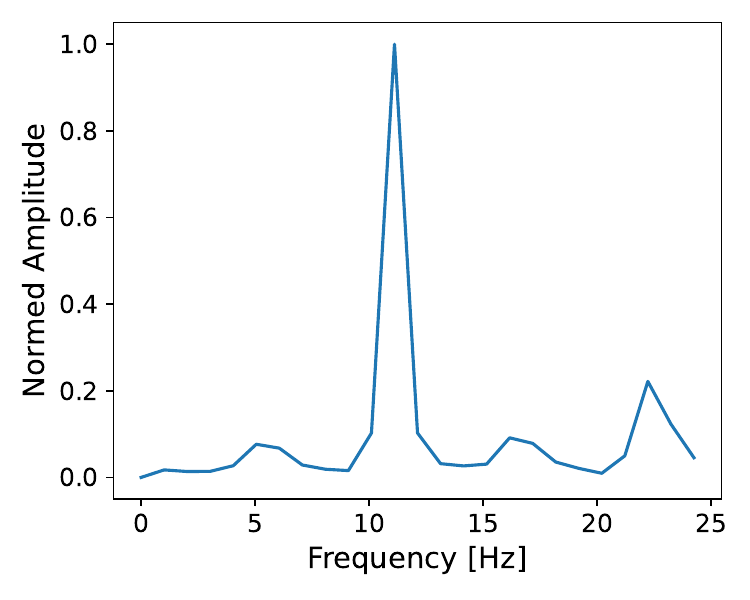}}
        \\
        \rotatebox{90}{\makecell{20.0 Hz}}
		&{\includegraphics[width=\linewidth]{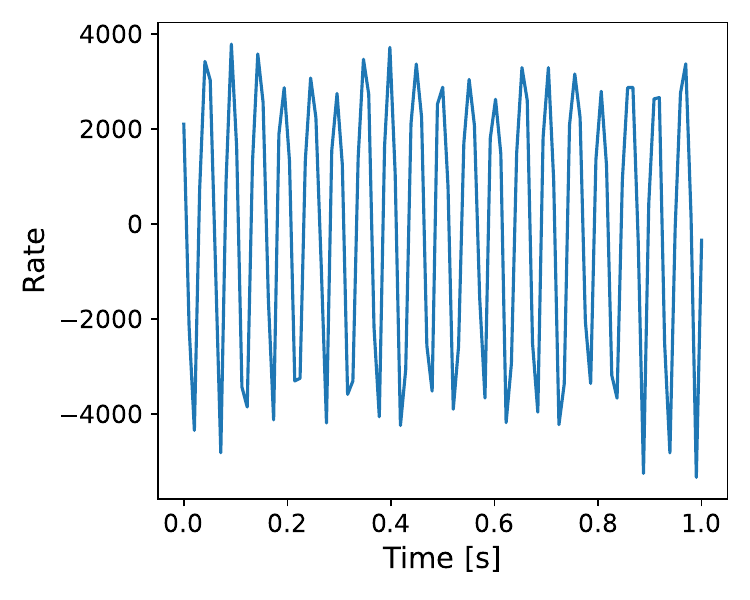}}
        &{\includegraphics[width=\linewidth]{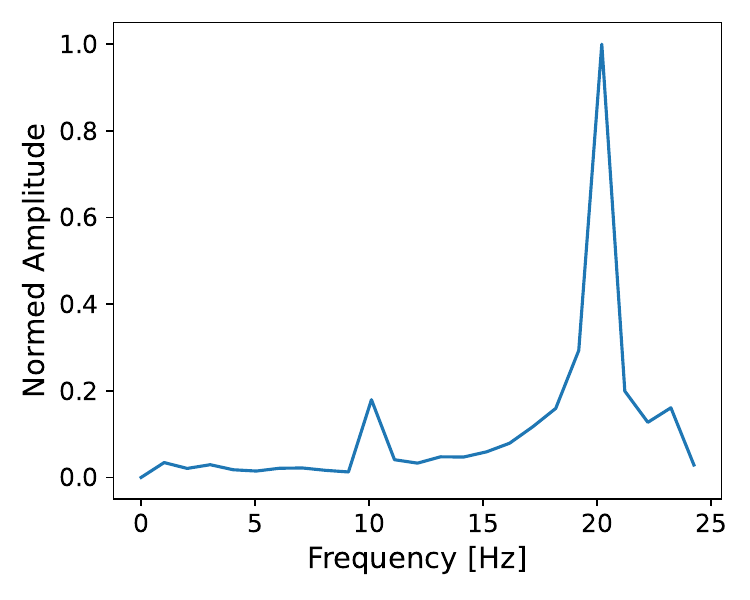}}
        \\

		& (c) Event rate
        & (d) Spectrum
	\end{tabular}
	}
	\caption{Rates and Fourier Spectra for the sequence ``4-bar\_test1'' from the EB\_SLAM dataset \cite{Chamorro20bmvc}. 
 The sequence contains data of a motor spinning with increasing frequency. 
 The clear peaks in the Fourier spectra indicate that our method could be extended to multi-class cases and for applications in vibration monitoring.
    }
	\label{fig:disc_eb_slam}
\end{figure}

The proposed system advances the measurement capabilities compared to previous technology, such as camera traps (\cref{sec:related_work}), 
which capture frames at low rate (e.g., 1 image per minute for long-term studies \cite{Morandini21pb})
or focused recordings (of 15-minute duration for short-term observation studies) \cite{Morandini21pb}. 
Both can miss EDs (since EDs last between 2 and 40 seconds \cite{Hamann24cvpr}, and they can happen throughout day and night). 
Even if a conventional-camera system was deployed for long-term observation of EDs, it would suffer from shortcomings such as large amount of temporally-redundant video data (which is undesirable for oscillatory motion analysis) and poor information in low-light or HDR conditions (increasing the exposure time would simply yield motion-blurred images).
By contrast, event data can pinpoint intervals of interest more easily \cite{Hamann24cvpr} and does not suffer from motion blur in low-light conditions. 
Additionally, an important aspect of systems for behavior observation is their ease of use by researchers in ecology or similar fields.
Not only do they need tools that deliver good results, they also need tools that are easy to use.
The spectrogram in \cref{fig:exp:spectrogram} or the spectra in \cref{fig:exp:qualitativepenguinbehaviors} illustrate how the different behaviors are distinguishable in the Fourier domain.
Beyond the use of the transformed signal for automated behavior classification, we found spectrograms to be a useful tool for initial data exploration.
They are intuitive and require little programming knowledge in today's mature scientific computing libraries. 
Such plots, built by leveraging the unique motion-sensitive characteristics of event cameras, are a quick way to inspect possible interesting sub-parts in longer data sequences.

The estimation of oscillatory motion frequencies with event cameras has drawn interest beyond animal behavior observation.
For instance, \cite{Pfrommer22arxiv} demonstrated the estimation of per-pixel frequencies through imaging but did not address classification problems.
In contrast, our Fourier-based approaches could also be used for monitoring machines and classifying vibrations. 
This is illustrated by an analysis of data from the EB\_SLAM dataset \cite{Chamorro20bmvc}, which contains two sequences of a motor spinning with increasing frequency.
The plots of the spectra in \cref{fig:disc_eb_slam} reveal clear peaks in the Fourier domain, which suggest that our scheme could be easily applied to industrial use-cases like vibration-based machine condition monitoring \cite{Woongjae23meas}. 
\begin{figure}[t]
	\centering
    {
    \setlength{\tabcolsep}{2pt}
	\begin{tabular}{
    >{\centering\arraybackslash}m{0.05\linewidth} 
	>{\centering\arraybackslash}m{0.45\linewidth} 
	>{\centering\arraybackslash}m{0.45\linewidth}}

        \rotatebox{90}{\makecell{Object}}
		&{\includegraphics[width=\linewidth]{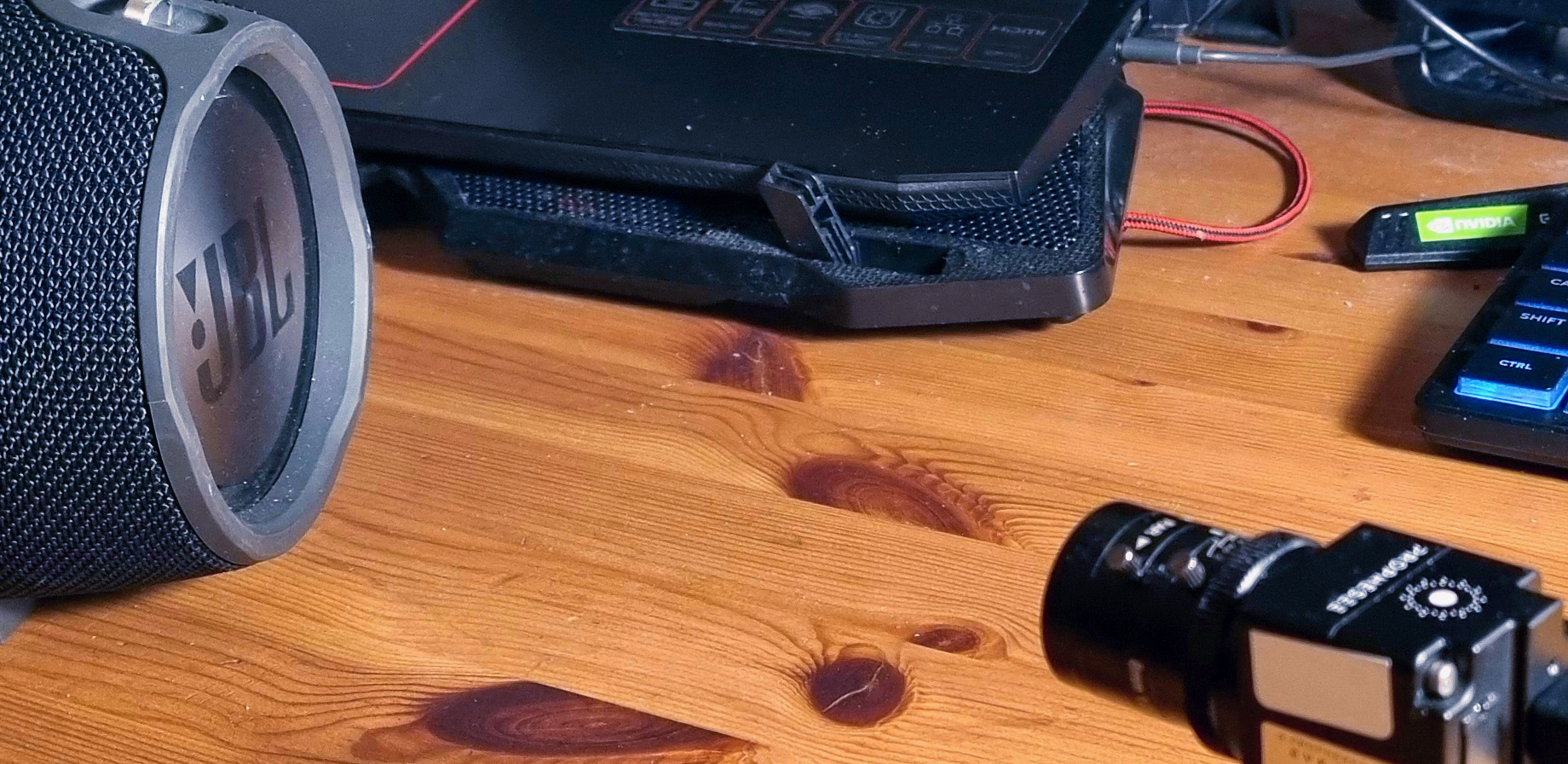}}
        &{\includegraphics[width=\linewidth]{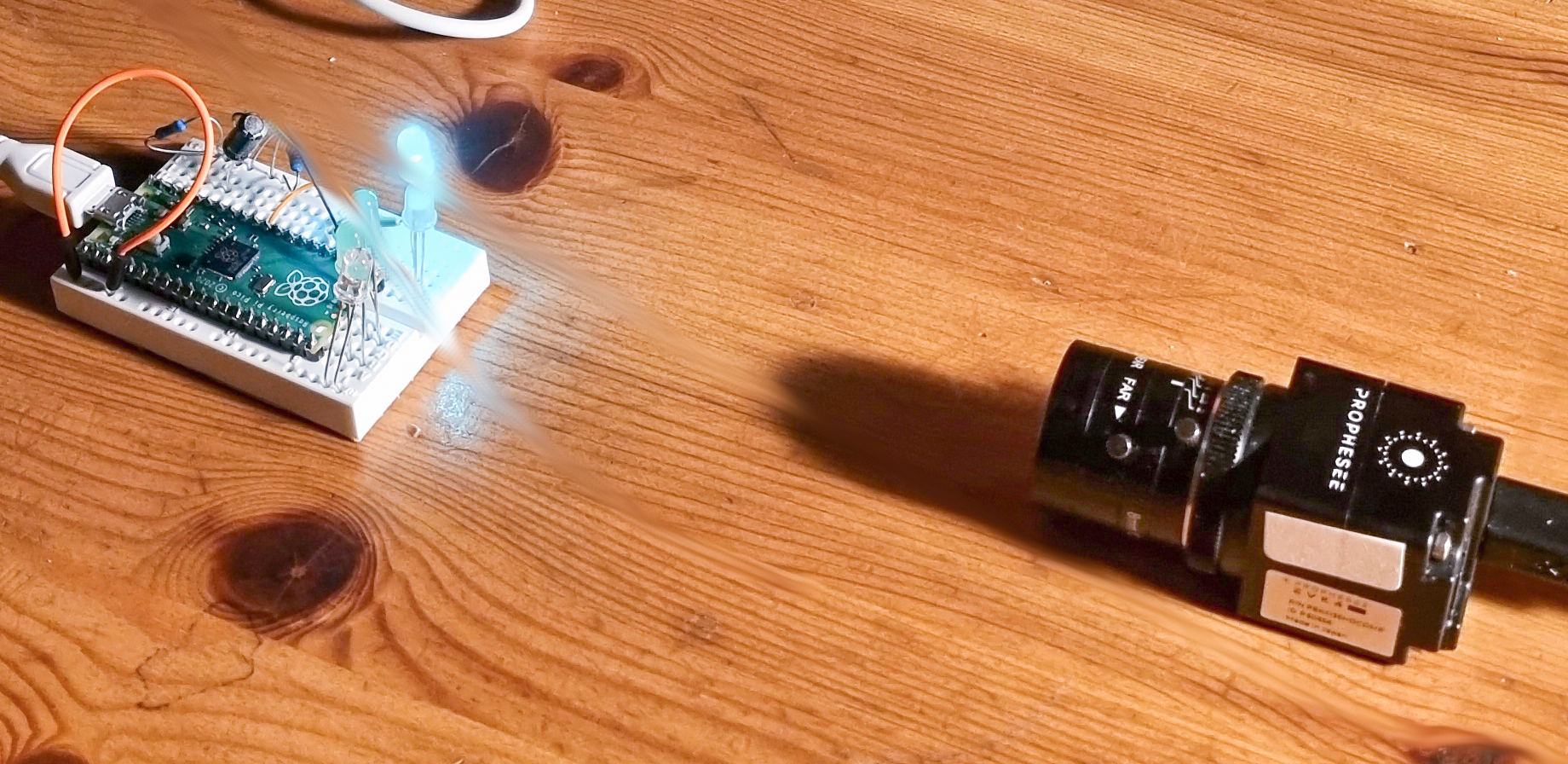}}
        \\
        \rotatebox{90}{\makecell{Events}}
		&\gframe{\includegraphics[width=\linewidth]{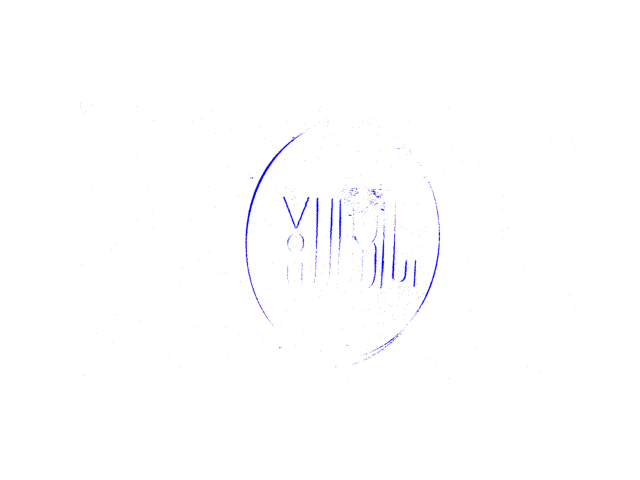}}
        &\gframe{\includegraphics[width=\linewidth]{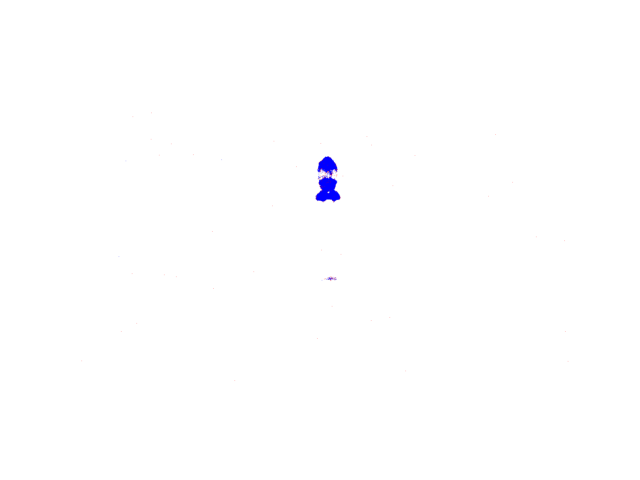}}
        \\
        \rotatebox{90}{\makecell{Spectrum}}
		&{\includegraphics[width=\linewidth]{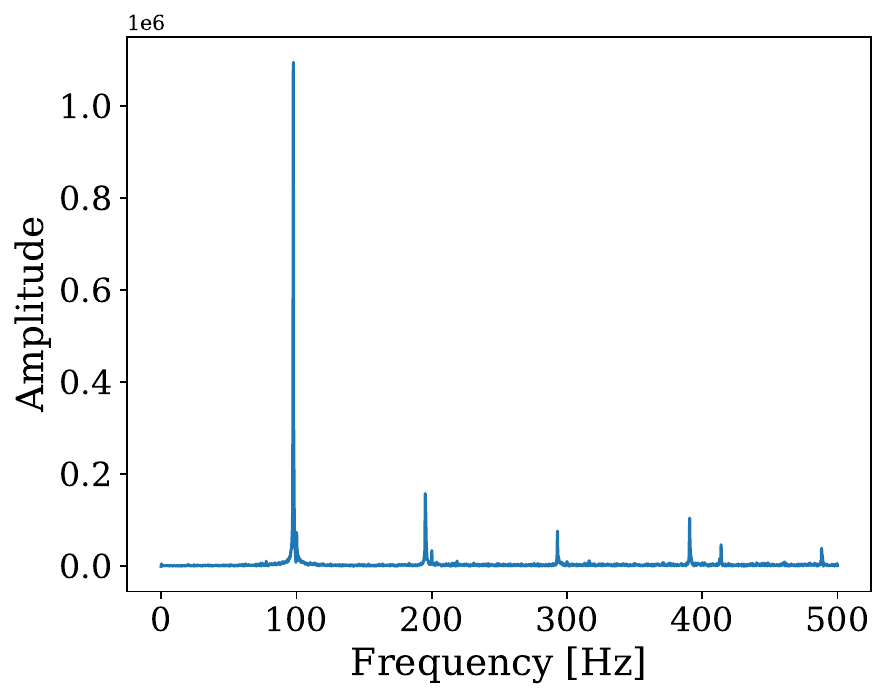}}
        &{\includegraphics[width=\linewidth]{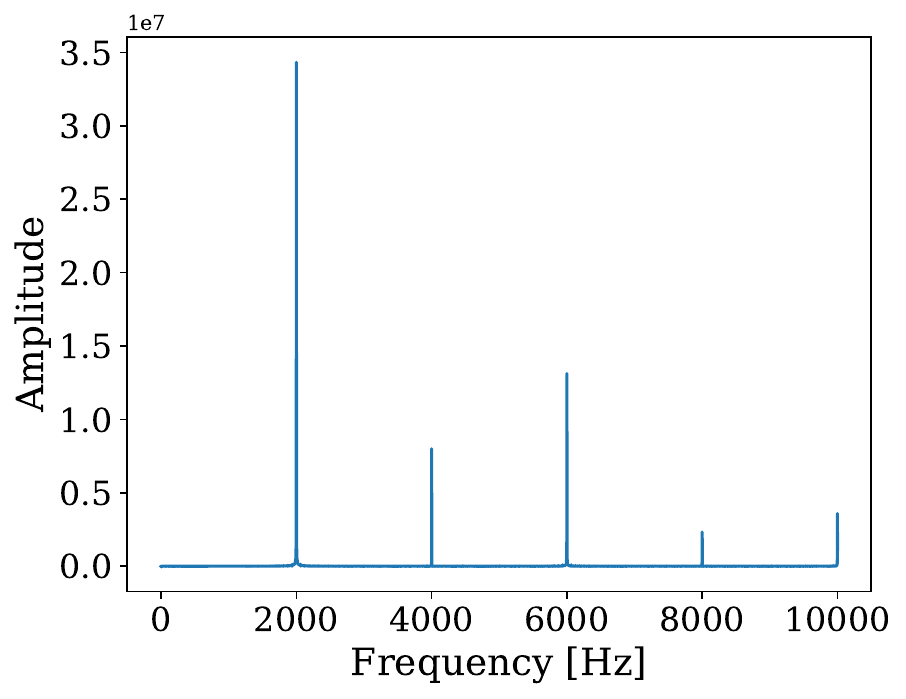}}
        \\

		& (a) Speaker, 100Hz
        & (b) Flashing LED, 2000Hz
	\end{tabular}
	}
	\caption{Event images and Fourier spectra of the EE3P dataset~\cite{Kolaf24cvww}.
 The examples show high-frequency phenomena, far beyond the framerates of conventional cameras.
 Our event-based method can be adapted to those use cases, where conventional cameras fail.
    }
	\label{fig:disc_ee3p}
\end{figure}

Our presented methods build on the advantages of event cameras, showing a way to solve the classification without relying on large artificial neural networks.
The dataset \cite{Hamann24cvpr} provides high-speed frame-based data for a direct comparison; 
however, \cref{fig:disc_ee3p} shows the first part of our pipeline on data from the EE3P (Event-based Estimation of Periodic Phenomena Properties) dataset~\cite{Kolaf24cvww}, containing event data for several high-frequency applications, e.g. LED blinking at 2000~Hz.
The visualized spectra show clear peaks at the respective working frequencies, suggesting that our method would work well for these examples.

\section{Conclusion} 
\label{sec:conclusion}

We tackled the task of recognizing oscillatory actions with event cameras.
We proved the feasibility of the proposed methods on a dataset of ecstatic displays, a unique behavior where a penguin flaps its wings, resulting in oscillating patterns in the event data.
The introduced methods show robust accuracy while relying on succinct classifiers.
The methods leverage the unique structure of the event data and yield simple data processing methods, which indicates that the event camera is a strong sensory match for this task. 
We focused on presenting a lightweight tool that can run in continuous time (as opposed to previously available remote observation technology), 
and showed its potential applicability to other problems of detecting oscillatory patterns in nature,
while the knowledge advancement in terms of penguin ecological insights is currently under investigation.

Building classifiers that do not rely on large neural networks could enable the development of online algorithms.
These lightweight algorithms could leverage the promising low-power properties of event cameras,
which in turn could open possibilities for edge-computing applications and facilitate the use of event cameras in long-term monitoring of animals \cite{Tuia22natcomm}, 
as well as in industry applications like vibration monitoring for condition-based maintenance of machines.

\section*{Acknowledgment}
This work was funded by the Deutsche Forschungsgemeinschaft (DFG, German Research Foundation) under Germany’s Excellence Strategy -- EXC 2002/1 ``Science of Intelligence'' -- project number 390523135.
This work was supported by the Oxford Berlin Research Partnership that is funded under the Excellence Strategy of the Federal Government and the L\"ander by the Berlin University Alliance, and by the University of Oxford. 

{\small
\bibliographystyle{IEEEtran}

}

\end{document}